\documentclass[preprint,12pt]{elsarticle}

\usepackage{amssymb}
\usepackage{amsmath}
\usepackage{hyperref}
\usepackage{caption}
\usepackage{float}
\usepackage{verbatim} 
\usepackage{booktabs, multirow}
\usepackage{amsfonts}
\restylefloat{figure}
\restylefloat{table}
\usepackage{graphicx}
\usepackage{adjustbox}
\usepackage{subcaption}
\usepackage{algorithm}
\usepackage{cleveref}
\usepackage{algpseudocode}

\journal{}

\begin{document}
	
	\begin{frontmatter}
		
		
		
		\title{Self-Supervised Auxiliary Task Discovery for Stable Reinforcement Learning in Stock Trading}

\author[label1]{Arishi Orra \corref{cor1}}
\ead{d21022@students.iitmandi.ac.in}  
\cortext[cor1]{Corresponding author}

\author[label1]{Himanshu Choudhary}
\ead{d21024@students.iitmandi.ac.in}

\author[label1]{Manoj Thakur}
\ead{manoj@iitmandi.ac.in}

\affiliation[label1]{organization={School of Mathematical and Statistical Sciences},
	addressline={Indian Institute of Technology Mandi}, 
	city={Mandi},
	postcode={175005}, 
	state={Himachal Pradesh},
	country={India}}

		\begin{abstract}
			Reinforcement learning has gained increasing attention as a data-driven approach for stock trading. However, learning a policy that is both profitable and stable remains challenging due to non-stationary market behaviour and noisy reward signals. Auxiliary tasks are often used to improve representation learning and stabilize training, yet they are usually designed manually and depend heavily on prior assumptions about targets and prediction horizons. Such fixed designs may not remain suitable across changing market regimes. In this work, we propose a self-supervised framework that automatically discovers auxiliary tasks to support reinforcement learning for stock trading. The auxiliary tasks are formulated as General Value Functions so that their predictions enrich the learned state representation and assist policy optimization. The framework consists of two networks. The main network learns the trading policy along with the auxiliary predictions, while the secondary network generates the definitions of auxiliary tasks through learned cumulants and discount factors. These tasks are updated using a meta gradient mechanism that accounts for their long-term impact on trading performance and improves training stability. We evaluate the proposed approach across four major equity indices: DJI, FTSE, Sensex, and TAIEX. The empirical results demonstrate that automatically discovered auxiliary tasks lead to more robust learning and improved trading performance compared to existing baselines.
		\end{abstract}
		
		
		
		\begin{keyword}
			Auxiliary Tasks \sep Deep Reinforcement Learning \sep Proximal Policy Optimization \sep General Value Functions \sep Stock Trading.
		\end{keyword}
		
	\end{frontmatter}
	
\section{Introduction} \label{Introduction}

Over the past two decades, automated stock trading has steadily transformed the global financial landscape. Equity markets generate a large volume of high-frequency data every day. Identifying profitable trading opportunities in this complex environment far exceeds the cognitive capabilities of individual human traders. Consequently, institutional and retail traders have widely started adopting algorithmic trading systems. These systems make decisions with minimal human involvement, operating at speeds far beyond human capabilities. These systems eradicate the emotional bias from trading and allow for trading large volumes of trades within seconds. However, the dynamic, non-stationary, and often noisy nature of financial markets presents persistent challenges for the design of robust automated trading agents. Due to the high volatility and complex interactions between trader psychology, global events, and asset prices, automated trading is a challenging technical problem \cite{ansari2022deep, li2022stock}.

Initial attempts to automate stock trading heavily relied on supervised learning, which focuses on predicting price or trend using historical data. Techniques such as linear regression, decision trees, support vector machines, and more recently, deep neural networks have been applied to forecast stock prices and trends \cite{pricope2021deep}. These models have demonstrated certain success when the market regime is stable and sufficient labeled data is available. In particular, deep learning models have shown enhanced predictive capability due to their ability to extract temporal and spatial features from raw price data. Nevertheless, supervised learning methods exhibit several fundamental drawbacks when applied to trading tasks. Firstly, they rely heavily on the statistical assumption that data remain stationary over time, which is rarely true in real markets. Second, the primary objective of most supervised models is to minimize forecast error, overlooking the goal of maximizing the trading profit. Also, they do not consider the most practical constraints, such as transaction costs, slippage, and liquidity constraints \cite{meng2019reinforcement}. As a result, supervised models perform well in backtests but often fail in live markets.

To overcome these limitations, researchers have formulated automated trading as a sequential decision-making problem well-suited to reinforcement learning (RL) paradigms \cite{choudhary2026cvar}. In the RL framework, the trading process is modeled as a Markov Decision Process (MDP) where an agent interacts with its environment at discrete time steps \cite{zhang2019deep, huang2023algorithmic}. At each step, the agent observes a market state, selects an action (buy, sell, or hold), and receives numerical feedback (profit or loss) corresponding to its trade \cite{shi2021stock, kabbani2022deep}. The agent learns to make optimal trading decisions through these repeated interactions, aiming to maximize expected cumulative rewards over time. This paradigm aligns perfectly with realistic trading, where actions are independent and the focus is on long-term success. Other financial domain problems like multi-asset trading, dynamic asset allocation, and portfolio rebalancing are also naturally expressible within the RL paradigm \cite{yang2023deep, xiong2018practical, orra2025deep}. Unlike static predictors, RL agents are capable of adapting dynamic strategies based on market regimes. RL agents can also incorporate risk measures like the Sharpe ratio or the drawdown by optimizing them as a long-term objective. Reinforcement learning combined with deep learning, i.e., deep reinforcement learning (DRL), leverages the representation power of neural networks to represent complex and high-dimensional state-action spaces \cite{li2022stock, kabbani2022deep}. In recent studies, practitioners have widely adopted modern DRL algorithms like A2C \citep{yu2023dynamic}, DDPG \cite{zhu2026rewards}, and PPO \citep{avramelou2024deep, orra2025enhancing} for the automated stock trading problem.

In the conventional RL setup, the agent's sole objective is to maximize a single extrinsic reward, which frequently results in low learning efficiency, particularly when feedback signals are delayed or infrequent \cite{choudhary2025risk}. Auxiliary tasks are secondary tasks alongside the main objective that help to improve the agent's performance. They guide the agent's learning process by providing additional pseudo-objectives derived from the environment itself \cite{veeriah2019discovery}. By learning the answers to them, the agent learns a richer representation of its environment, improves data efficiency, and accelerates learning on the main task. Some of the most commonly used auxiliary tasks are pixel control, next state prediction, reward prediction, and feature control \cite{jaderberg2016reinforcement}. Even in stock trading, the agent often encounters sparse and highly volatile rewards due to the noisy market conditions. Such reward signals can slow the learning process and lead to poor generalizability. Auxiliary tasks such as volatility estimation \cite{sun2022deepscalper}, price change prediction \cite{konstantinos2022reinforcement}, and self-supervised tasks \cite{arabha2024improving} have been recently adopted by researchers to improve the efficacy of the agent. By jointly optimizing these secondary tasks along with the main task, the agent develops richer representations of the market state and improves the overall profitability. 

Designing handcrafted auxiliary tasks for stock trading is inherently challenging due to the dynamic and complex nature of financial markets. An auxiliary task that performs well under certain market conditions may fail under others, highlighting a lack of robustness and generalization. Such tasks often require significant domain knowledge, making them difficult to scale across diverse and evolving market environments. To address these limitations, we present QUESTrader (Question Enhanced Stock Trader), a novel method for automatically discovering auxiliary-task questions in stock trading environments. The proposed framework encodes auxiliary questions as General Value Functions specified by cumulants and discounts. QUESTrader adopts a dual-network design, where the main network learns the policy, value function, and GVF answers, and the question network proposes the auxiliary questions by emitting cumulants and discounts. We then introduce a meta-gradient approach that treats the question network parameters as meta-parameters and utilizes a non-myopic meta-gradient RL approach to refine the questions. It assigns credit to auxiliary questions for their delayed impact on trading performance and enhances the training stability. The effectiveness of the proposed QUESTrader framework is established through empirical evaluation across four prominent stock market indices: DJI, Sensex, FTSE, and TAIEX. The performance of the proposed method is benchmarked against baseline trading strategies across multiple metrics capturing both risk and return characteristics. The experimental results on these datasets validate the superiority of the proposed model, with consistent advantages observed in the evaluation metrics, substantiated by the ablation studies.

In summary, our main contributions can be outlined as follows:
\begin{enumerate}
	\item We propose QUESTrader, a novel framework for automatically discovering auxiliary-task questions for stock trading. It employs a two-network architecture: the answer (main) network learns the policy and the answers to the generated questions, while the question network formulates these questions as GVFs.
	\item We introduce a non-myopic meta-gradient approach that treats the question network parameters as meta-parameters. It assigns credit to auxiliary questions for their delayed impact and enhances training stability.
	\item We conduct extensive empirical evaluation on four major global equity indices: DJI, FTSE, Sensex, and TAIEX. The experimental findings reveal that QUESTrader consistently achieves higher returns and superior risk-adjusted ratios with competitive drawdowns compared to strong baselines.
	\item We also conducted a detailed ablation study to analyze the influence of key hyperparameters, including the number of discovered GVF questions and the inner unroll length, providing insight into how auxiliary questions shape representations and affect trading performance.
\end{enumerate}

\section{Related Work} \label{Literature Review}

\subsection{RL for stock trading}

In recent years, RL methods have become increasingly prevalent in automated stock trading applications. Moody et al. \cite{moody1998performance} were the pioneers in exploring the application of RL in stock trading. They proposed a recurrent RL (RRL) model for trading optimized with a novel metric differential Sharpe ratio. Empirical studies demonstrated the effectiveness of RL in quantitative trading by outperforming the traditional supervised learning methods. As an extension to \cite{moody1998performance}, Dempster and Leemans \cite{dempster2006automated} proposed an adaptive RRL framework for forex trading. The proposed system employs a three-layer architecture, utilizing RRL as the base algorithm. Gao and Chan \cite{gao2000algorithm} proposed QSR, a trading agent that utilizes Q-learning and the Sharpe ratio maximization algorithm. It is trained by maximizing the absolute profit and relative risk-adjusted profit, and employs a combination of two networks for testing. With advancements in deep learning, various DRL approaches have been introduced for algorithmic trading. Deng et al. \cite{deng2016deep} enhanced the traditional RRL model by using an RNN layer to extract meaningful temporal representations of the market. Also, they introduced a fuzzy extension to counter the uncertainty in the data. Lu \cite{lu2017agent} adopted the RRL technique, utilizing policy gradient combined with an LSTM network to capture sequential dependencies effectively. To handle the noisy market data, GRU is employed by Wu et al. \cite{wu2020adaptive} to extract informative features from the raw data. These financial features are then integrated into DQN and DDPG models to develop adaptive trading strategies. As an alternative to existing methodologies, Kabbani and Duman \cite{kabbani2022deep} reformulated the stock trading task as a Partially Observed MDP (POMDP) and implemented the TD3 algorithm for decision-making in trading. In a recent study, Qin et al. \cite{qin2024earnhft} proposed EarnHFT, a hierarchical RL framework for high-frequency trading. It first trains low-level agents for different market regimes, and then the high-level agents dynamically select the most suitable lower agent from the pool. Huang et al. \cite{huang2024novel} presented efficient deep SARSA, a novel DRL framework integrating bidirectional LSTM with an attention mechanism for trading in volatile markets. Orra et al. \cite{orra2024dynamic} proposed Dynamic Reinforced Ensemble with Bayesian Optimization (DREB), an ensemble approach that dynamically adapts to varying market conditions. Bayesian optimization is utilized to dynamically assign time-varying weights to various base DRL models.

\subsection{Auxiliary tasks in financial trading}

Auxiliary tasks are the secondary objectives that help the agent to learn a better market representation and increase the overall performance. Sun et al. \cite{sun2022deepscalper} proposed DeepScalper, a risk-aware RL framework for intraday trading. In addition to their primary objective, they included a risk-aware auxiliary task of predicting the market volatility. This additional task helped the agent identify the market risk while maximizing the profit. In his thesis work, Konstantinos \cite{konstantinos2022reinforcement} enhanced the performance of the PPO model by adding the auxiliary task of predicting the price movement. The empirical findings demonstrated that adding this additional task stabilized training and improved the overall performance. Instead of using a single auxiliary task, Ong and Herremans \cite{ong2023constructing} utilized multiple secondary tasks related to volatility forecasting. The results indicated that adding volatility-based objectives provided superior risk-adjusted returns and robustness. Arabha et al. \cite{arabha2024improving} improved the performance of the PPO model by utilizing an auxiliary task derived from self-supervised learning logic for Forex trading. The agent is tasked to classify patterns and extract additional information from the Forex data. The experimental results confirmed the effectiveness of adding the auxiliary task in increasing the overall return. To handle the noisy and nonstationary data, Sang et al. \cite{sang2025portfolio} proposed a multi-task self-supervised learning module that enriches the market representations by generating auxiliary tasks. They formulated these secondary tasks of predicting the asset trends and volatility into binary classification problems. This formulation improved the model's profitability and robustness by extracting better feature representations.

\section{Background \label{Background}}

This section lays the technical foundations used throughout the paper. It reviews the fundamentals of reinforcement learning, outlines the PPO algorithm, and introduces the concept of GVFs. Finally, it formalizes the stock trading problem within the MDP framework by specifying the state, action, and reward.

\subsection{Reinforcement Learning}

Reinforcement learning (RL) is a robust machine learning framework for solving sequential decision-making problems. In RL, an agent learns to make decisions by interacting with its environment through trial and error. The agent balances the exploration of new strategies with the exploitation of known good actions, aiming to maximize cumulative rewards. Formally, the RL environment is modeled as a Markov decision process (MDP): $(S, A, P,r,\gamma)$ \cite{kaelbling1996reinforcement}, where 
\begin{itemize}
	\item $S$  is the set of possible states.
	\item $A$ is the set of possible actions.
	\item $\mathbb{P}:S \times A \rightarrow S$ is the state transition function, which determines the probability distribution over next states given the current state and action.
	\item $r: S \times A \rightarrow \mathbb{R}$ is the reward function, specifying the immediate reward received after performing an action in a given state.
	\item $\gamma \in [0,1]$ is the discount factor, determining the present value of future rewards.
\end{itemize}  
A policy $\pi: S \rightarrow A$ is a mapping from states to a probability distribution over actions. At each discrete time step $t$, the agent observes a state $s_t \in S$, takes an action $a_t \in A$, receives a numerical feedback reward $r_t(s_t,a_t)$, and then transitions to a new state $s_{t+1}$ with probability $\mathbb{P}(s_{t+1} |s_t,a_t)$. The goal in RL is to find an optimal policy $\pi^*$ that maximizes the expected sum of discounted rewards, i.e., $\mathbb{E}_\pi \left[ \sum_{t=0}^\infty \gamma^t r_t \right] $ \cite{sutton1998reinforcement}.

To evaluate how good it is to be in a particular state, RL introduces the state value function of a policy $\pi$ as:
\begin{equation}
	V^\pi(s) = \mathbb{E} \left[ \sum_{t=0}^\infty \gamma^t r_t | s_0 =s \right].
\end{equation}
It measures the total expected discounted reward starting from state $s$ and thereafter following the policy $\pi$. Similarly, the action value function or Q-function is the expected return when starting from state $s$, taking action $a$, and thereafter following the policy $\pi$:
\begin{equation}
	Q^\pi(s,a) = \mathbb{E} \left[ \sum_{t=0}^\infty \gamma^t r_t | s_0 =s, a_0=a \right].
\end{equation}
The optimal value function $V^*(s)$ or optimal Q-function $Q^*(s,a)$ satisfies the Bellman optimality equation:
\begin{equation}
	Q^*(s,a) = \mathbb{E} \left[ r + \gamma \max_{a'} Q^*(s',a')|s,a \right]
\end{equation}
where $s'$ and $a'$ are the next state and action, respectively \cite{wang2021improving}. 

The value-based methods, such as Q-learning, learn to approximate the value function or Q-function using parameterized models \cite{mnih2013playing}. The policy is then derived by selecting the actions that yield the maximum value, i.e.,
\begin{equation}
	\pi^*(s_t) = \underset{a}{\mathrm{argmax}} \quad Q^*(s_t,a).
\end{equation}
In contrast, the policy-based methods directly approximate the optimal policy by maximizing the expected reward through gradient-based optimization \cite{sutton1999policy}. It learns a policy parameterized by $\theta$ by maximizing:
\begin{equation}
	J(\theta) = \mathbb{E}_\pi \left[ \pi_\theta (a|s) \hat{A}^\pi(s,a) \right]
\end{equation}
where $\hat{A}^\pi(s, a) = Q^\pi(s, a) - V^\pi (s)$ is the advantage function that quantifies the gain from taking action $a$ in state $s$ compared to the average action. Another class of methods is the actor-critic method, which combines ideas from value-based and policy-based methods. It uses an actor network to estimate the policy and another critic network to approximate the value function \cite{mnih2016asynchronous}. The critic evaluates the actions taken by the actor and provides feedback in the form of value estimates, which are then used to update the policy in a more stable and data-efficient manner.

\subsection{Proximal Policy Optimization}

Proximal Policy Optimization (PPO) has recently emerged as one of the most stable and effective methods in deep RL, widely used in training large-scale systems like financial trading agents and language models \cite{tsantekidis2020price, del2024comparative}. Policy gradient methods often incur the issue of training instability due to large policy updates. Trust Region Policy Optimization (TRPO) improved stability by limiting the policy updates within a fixed trust region, but it introduced significant complexity. PPO simplifies TRPO by enforcing a softer constraint, making it practical, stable, and easier to implement \cite{del2024comparative}.

PPO employs two neural networks: the policy network (Actor) and the value network (Critic). The policy network, parameterized by $\theta$, maps states to actions, whereas the value network, parameterized by $\phi$, predicts the expected cumulative reward from given states. The core strength of PPO lies in its clipped surrogate objective, ensuring stable policy updates by preventing drastic shifts in behavior between successive policy iterations. The PPO objective function can be expressed as:

\begin{equation}
	J_{\text{CLIP}}(\theta) = \mathbb{E}_t\left[ \min \left(\hat{r}_t(\theta)\hat{A}_t, \, \text{clip}(\hat{r}_t(\theta), 1 - \epsilon, 1 + \epsilon)\hat{A}_t\right)\right],
\end{equation}

where the probability ratio \(\hat{r}_t(\theta)\) compares the likelihood of selecting an action \(a_t\) at a state \(s_t\) under the new policy parameters \(\theta\) against the old parameters \(\theta_{\text{old}}\):

\begin{equation}
	\hat{r}_t(\theta) = \frac{\pi_{\theta}(a_t|s_t)}{\pi_{\theta_{\text{old}}}(a_t|s_t)}.
\end{equation}

The function \(\text{clip}(\hat{r}_t(\theta), 1 - \epsilon, 1 + \epsilon)\) restricts the policy updates within a narrow interval defined by the hyperparameter \(\epsilon\). This clipping effectively prevents excessive updates, ensuring that the new policy remains relatively close to the previous one, thereby enhancing the stability and robustness of the training process \cite{schulman2017proximal}. Additionally, the value network $V_\phi$ is trained to minimize the temporal-difference error using mean squared error:
\begin{equation}
	J_{critic}(\phi) = \mathbb{E}_t \left[ ||V_\phi(s_t) - \hat{R}_t||^2 \right].
\end{equation}
Here, $\hat{R}_t$ is the actual return value for state $s_t$ and is computed as $\hat{R}_t = \sum_{l=0}^\infty \gamma^l r_{t+l}$.

\subsection{General Value Functions}

General Value Functions (GVFs) extend the concept of traditional value functions to represent richer forms of knowledge beyond just reward prediction. Instead of predicting future rewards alone, GVFs generalize this concept by predicting expected cumulative values of any signals or features of the environment. Formally, a GVF is defined as the expectation of a cumulant signal discounted by a state-dependent discount factor, following a certain policy. Mathematically, GVFs can be represented as:
\begin{equation}
	V^{\pi, \gamma, c}(s) = \mathbb{E}_\pi \left[ \sum_{k=t}^\infty \left( \prod_{i=t+1}^k \gamma(s_i) \right) c_{k+1} \mid S_t=s, A_{t:\infty} \sim \pi\right] 
\end{equation}
where $c(\cdot)$ is the cumulant and $\gamma(\cdot)$ is the state-dependent discount factor \cite{sutton1998reinforcement}. Unlike the conventional value functions, GVFs allow these cumulants to be arbitrary, effectively enabling a wide variety of questions or predictions about the environment. GVFs have emerged as powerful auxiliary tasks to enhance the learning efficiency of RL agents. By discovering GVFs through meta-gradient methods, agents can autonomously generate and refine auxiliary questions tailored to support representation learning relevant to their main objectives \cite{veeriah2019discovery}.

\subsection{MDP Formulation for Stock Trading}

We formulated the multi-stock trading problem as an MDP. Each outcome of the trading decision depends on both the current market state and the actions taken in previous periods. Thus, due to the sequential decision-making nature, it fits naturally into the framework of MDP. Modeling stock trading as an MDP enables the use of RL to optimize strategies over time.

The agent's objective is to optimally allocate the available trading capital by simultaneously executing trades across all the stocks. Let the trading horizon be discretized into $T$ periods, indexed by $ t=1,\ldots, T$. At each time $t$, the agent observes the market state $s_t$ and executes an action $a_t \in \{ buy, sell, hold\}$. As a result of its action, the agent receives a reward $r_t$ (profit/loss) from the environment and transitions to the next state $s_{t+1}$. The goal of the agent is to determine a policy $\pi$ that maximizes the expected discounted cumulative rewards, i.e., 
\begin{equation}
	G_t = \mathbb{E}_\pi \left[ \displaystyle \sum_{t=0}^{\infty} \gamma^t r(s_t, a_{t}) \right].
\end{equation} 
Optimizing $G_t$ aligns precisely with our goal of maximizing the expected cumulative wealth \cite{zhang2019deep}. In our problem, $S,A$, and $r$ are set as follows.

\paragraph{State Space} Each state $s_t \in S$ captures all the information required for decision-making at time $t$.  For a $n-$stock problem, each state $s_t$ is a $(10n+1)$ dimensional vector. A list of components of the state vector is below:
\begin{itemize}
	\item Remaining balance available with the agent.
	\item Number of shares of each stock that the agent holds.
	\item Close price of each stock.
	\item Eight technical indicators ($30$ and $60$ day Simple Moving Averages (SMA), Moving Average Convergence Divergence (MACD), upper and lower Bollinger bands, Relative Strength Index (RSI), Commodity Channel Index (CCI), and Average Directional Index (ADX)) (\cite{murphy1999technical}) corresponding to each stock.
\end{itemize}

\paragraph{Action Space} The action $a_t \in A$ specifies the trading position in each stock at time period $t$. It is represented as a $n-$dimensional vector $a_t =[a_{t,1}, ..., a_{t,n}]^\top$, where $a_{t,i} \in \{-m, \ldots, 0,\ldots,m\}$ is the number of shares to be bought or sold in stock $i$ at time $t$. Here, $m$ is the maximum number of shares that can be traded at a particular time. Therefore, in our problem, the complete action space is of dimension $(2m+1)^n.$

\paragraph{Reward Function} The reward function quantifies the immediate benefit of each action and guides the agent to refine its policy. In this work, we are using the immediate change (profit or loss) in the account value as the immediate reward, i.e.,
$$
r(s_t,a_t) = (P_{t+1}-P_t)\cdot a_t - \delta P_t \cdot |a_t-a_{t-1}|
$$
where $P_t$ is the vector of closing prices of all stocks at time $t$ and $\delta$ is the fixed transaction cost. The first term in the reward is the instant profit or loss, and the second term is the transaction cost incurred after executing the trade $a_t$. This reward function incentivizes the agent to make profitable decisions while simultaneously minimizing transaction costs.

To achieve a realistic modeling scenario, several standard assumptions \cite{cui2024multi, jiang2017deep} are made:
\begin{enumerate}
	\item Zero Slippage: Sufficient market liquidity is assumed such that the trades are executed at the observed price.
	\item Negligible Market Impact: The trades do not influence the market prices.
	\item Immediate Settlement: Asset volume is large enough such that the settlement occurs at the end of each period.
\end{enumerate}

\section{Methodology \label{Methodology}}

Jointly learning auxiliary tasks with the trading policy enhances representation learning, highlights microstructure signals, and increases sample efficiency in market settings. Hand‑designing such tasks is challenging due to shifting market regimes, evolving objectives, and many candidate questions may misalign with the main objective. We therefore propose to discover the auxiliary questions automatically as General Value Functions, using a non‑myopic meta‑gradient that optimizes these questions to enhance downstream trading performance. Extending the principles of Veeriah et al. \cite{veeriah2019discovery}, we present QUESTrader, a novel framework that leverages question-enhanced auxiliary tasks to improve stock trading performance.

\subsection{Two‑Network Architecture for Discovery}

\begin{figure}[ht!]
	\centering
	\begin{subfigure}[b]{0.45\textwidth}
		\centering
		\includegraphics[width=\textwidth]{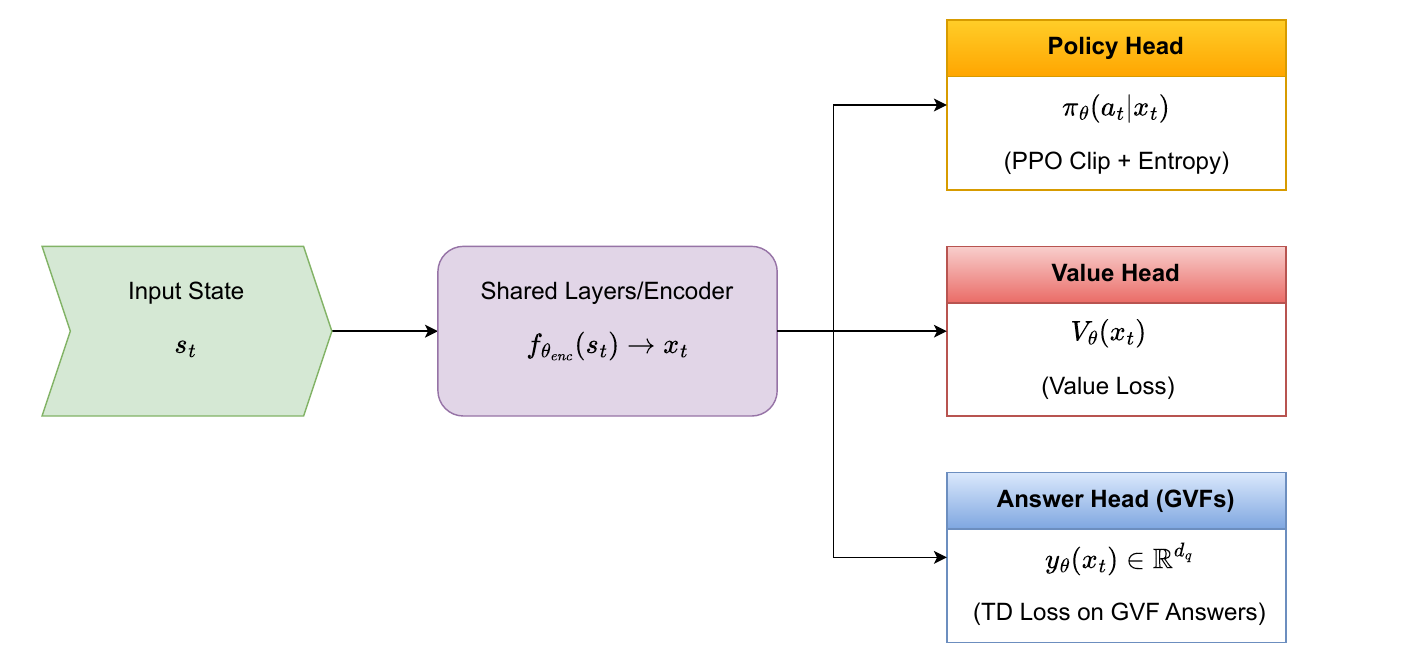}
		\caption{Main/Answer Network}
	\end{subfigure}
	\hfill
	\begin{subfigure}[b]{0.45\textwidth}
		\centering
		\includegraphics[width=\textwidth]{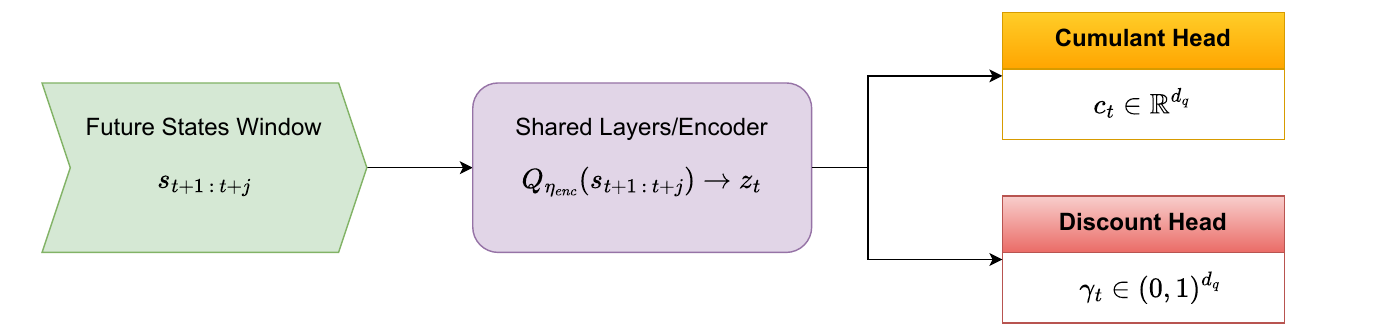}
		\caption{Question Network}
	\end{subfigure}
	\caption{The two-network architecture of the proposed QUESTrader.}
	\label{fig:network}
\end{figure}

We adopt a two‑network architecture for auxiliary‑task discovery in stock trading. The Answer/Main network learns the trading policy, value function, and the answers to auxiliary questions. The Question network proposes the auxiliary questions by emitting cumulants and discounts that define on‑policy GVFs. In this study, we adopt PPO as the underlying reinforcement learning algorithm for training the trading agent. An illustration of the architectural design of the two networks is presented in Figure \ref{fig:network}.

The question network, parameterized by $\eta$, receives a short future slice $s_{t+1:t+j}$ and produces two vectors $c_t \in \mathbb{R}^{d_q}$ and $\gamma_t \in \mathbb{R}^{d_q}$, where $d_q$ is the number of questions. Each pair $(c_t^{(i)}, \gamma_t^{(i)})$ specifies a GVF question. The question network's primary objective is to propose questions whose answers, when learned by the main network, yield representations that assist the PPO task. It has no direct supervised signal from the market. The question network is trained to minimize the loss:
\begin{equation}
	J_{\text{ques}} = J_{\text{PPO}}(\theta),
\end{equation}
where $J_{\text{PPO}}(\theta)$ is the standard PPO loss.

The main network, parameterized by $\theta$, receives the current state $s_t$ as input and produces the policy $\pi_\theta(a_t|s_t)$, the value function $V_\theta(s_t)$, and a set of scalar predictions $y_\theta(s_t) \in \mathbb{R}^{d_q}$ with one output corresponding to each discovered auxiliary question. The objective of the main network is twofold: to optimize trading performance under PPO and to accurately predict the answers to the discovered auxiliary questions. For question $i \in \{ 1, \ldots, d_q \}$, the answer head predicts:
\begin{equation}
	y_{\theta}^{(i)}(s_t) \approx \mathbb{E}_{\pi} \left[ \sum_{n \ge 0} \left( \prod_{m=0}^{n-1} \gamma_{t+m}^{(i)} \right) c_{t+n+1}^{(i)} \, \bigg| \, s_t \right],
\end{equation}
where $c^{(i)}$ and $\gamma^{(i)}$ come from the question network. Using a truncation horizon $W$, the TD target is:
\begin{equation}
	G_t^{(i)} = \sum_{n=0}^{W} \left( \prod_{m=0}^{n-1} \gamma_{t+m}^{(i)} \right) c_{t+n+1}^{(i)} 
	+ \left( \prod_{m=0}^{W} \gamma_{t+m}^{(i)} \right) y_{\theta'}^{(i)}(s_{t+W+1}).
\end{equation}

The \textit{answer loss} averages squared TD errors:
\begin{equation}
	J_{\text{ans}}(\theta; \eta) = \frac{1}{d_q} \sum_{i=1}^{d_q} \frac{1}{2} \mathbb{E} \left[ \left( G_t^{(i)} - y_{\theta}^{(i)}(x_t) \right)^2 \right].
\end{equation}

Then, the total loss of the main network is:
\begin{equation}\label{eq:main_loss}
	J_{\text{main}}(\theta; \eta) = J_{\text{PPO}}(\theta) + \lambda_{\text{aux}} J_{\text{ans}}(\theta; \eta),
\end{equation}
where $\lambda_{\text{aux}}$ is a weighting coefficient that controls the contribution of the auxiliary loss to the overall objective.

\subsection{Meta‑objective and Non‑Myopic Meta‑Gradient}

A meta‑gradient is the gradient of a meta‑loss with respect to meta‑parameters that shape how the learner itself learns. In this study, the meta‑parameters $\eta$ belong to the question network, which defines the auxiliary GVF questions. Changing $\eta$ changes the GVF targets, which changes how $\theta$ updates over many iterations; the meta‑gradient assigns credit to $\eta$ based on its long-term impact on PPO performance. The meta-gradient algorithm employs a bi-level optimization framework, wherein the inner loop performs $K$ updates to the main task parameters $\theta$, while the outer loop applies a single update to the meta parameters $\eta$. The step-by-step pseudo-code of the proposed method is detailed in Algorithm \ref{alg:QUESTrader}. 

\begin{algorithm}[H]
	\caption{PPO with Discovered GVF Aux Tasks (QUESTrader)}
	\label{alg:QUESTrader}
	\begin{algorithmic}[1]
		\Statex \textbf{Input:} $\theta_0, \eta_0$; inner step $\alpha$; meta step $\beta$; unroll $K$, PPO hyperparameters; $\lambda_{\text{aux}}$.
		\For{t $= 1$ to $N$}
		\State $\theta_{t,0} \leftarrow \theta_t$
		\State Compute $(c_t,\gamma_t)$ for GVF targets
		\For{$k = 0$ to $K$}
		\State Generate batches/trajectories under $\pi_{\theta_{t,k-1}}$
		\State Compute the main network loss $J_{\text{main}}(\theta_{t,k-1}; \eta_t)$ using equation \ref{eq:main_loss}
		\State $\theta_{t,k} \leftarrow \theta_{t,k-1} - \alpha \nabla_{\theta_{t,k-1}}J_{\text{main}}(\theta_{t,k-1}; \eta_t)  $
		\EndFor
		\State Compute meta-loss $M_t(\eta_t)$ using equation \ref{eq:meta_loss}
		\State $\eta_{t+1} \leftarrow \eta_t - \beta \nabla_{\eta_t} M_t(\eta_t)$
		\State $ \theta_{t+1} \leftarrow \theta_{t,K}$ 
		\EndFor
	\end{algorithmic}
\end{algorithm}

At each outer iteration $t$, the meta‑objective evaluates main‑task quality across the $K$ inner unroll and is defined as:
\begin{equation}\label{eq:meta_loss}
	M_t(\eta) = \sum_{k=1}^K J_{\text{PPO}}(\theta_{t,k}).
\end{equation}
The meta-loss may be computed after one update (myopic) or after $k>1$ updates (non-myopic), allowing for varying depths of credit assignment. The non-myopic meta-gradient differentiates $\mathcal{M}_t$ through the unrolled $K$ inner updates. Applying the chain rule across the trajectory $\theta_{t,0} \rightarrow \ldots \rightarrow \theta_{t,K}$ yields:
\begin{equation}
	\nabla_{\eta} \mathcal{M}_t(\eta) 
	= \sum_{k=1}^{K} 
	\underbrace{\frac{\partial J_{\text{PPO}}(\theta_{t,k})}{\partial \theta_{t,k}}}_{\text{main-task}}
	\underbrace{\frac{\partial \theta_{t,k}}{\partial \eta}}_{\text{how questions shape future learners}}.
\end{equation}
The meta-parameters $\eta$ are updated using a meta step of size $\beta$:
\begin{equation}
	\eta_{t+1} = \eta_t - \beta \nabla_{\eta} \mathcal{M}_t(\eta).
\end{equation}

\section{Experimental Setup \label{Experiments}}

This section describes the experimental setup used to evaluate QUESTrader. It outlines the datasets employed in the study, the evaluation metrics for assessing risk and return, the baseline methods for comparison, and the training and validation procedures.

\subsection{Dataset Description}

The effectiveness and robustness of the proposed methodology have been assessed using data from four prominent global stock market indices: the Dow Jones Industrial Average (DJI) from the US, the Financial Times Stock Exchange $100$ (FTSE) from the UK, the Sensex from India, and the Taiwan Capitalization Weighted Stock Index (TAIEX)  from Taiwan. DJI represents $30$ leading companies listed on major US stock exchanges, FTSE consists of the $100$ largest blue-chip companies by market capitalization traded on the London Stock Exchange, Sensex reflects a market-weighted index of the top $30$ companies listed on the Bombay Stock Exchange, and the TAIEX encompasses all publicly traded companies on the Taiwan Stock Exchange. To ensure consistency across all the experiments, we selected all 30 stocks from DJI and Sensex, and only the top 30 stocks from FTSE and TAIEX. For each index, we collected the daily closing price data from Yahoo Finance\footnote{\url{https://finance.yahoo.com/}}, spanning from January $1$, $2010$, to March $31$, $2025$. The closing price data is used to derive the technical indicators that act as state features. The complete dataset is partitioned into two non-overlapping subsets to facilitate model training and evaluation. Data from January $1$, $2010$, to December $31$, $2023$, is utilized for training and validating the models, while their out-of-sample trading performance is evaluated over the period from January $1$, $2024$, to March $31$, $2025$.

\subsection{Performance Metrics}

The effectiveness of the proposed approach in automated stock trading is evaluated using six widely accepted performance metrics. These metrics encompass return indicators, risk measurements, and risk-return tradeoff ratios.

\begin{enumerate}
	\item \textbf{Cumulative Return} (CR): Cumulative return measures the total change in the value of an investment over the entire investment horizon. It represents the aggregate growth (or decline) from the beginning to the end of the period. Mathematically,
	\begin{equation}
		CR = \frac{V_T}{V_0}-1
	\end{equation}
	where $V_0$ and $V_T$ are the initial and final investment values, respectively.
	
	\item \textbf{Annual Return} (AR): Annual return translates total investment growth into a consistent yearly percentage, accounting for compounding effects. For $N$ trading days, it can be calculated as:
	\begin{equation}
		AR = (1 + CR)^\frac{252}{N} - 1
	\end{equation}
	considering $252$ trading days as one year.
	
	\item \textbf{Sharpe Ratio} (SR): The Sharpe ratio is a risk-adjusted metric that evaluates how much excess return a trading strategy produces per unit of total risk. A higher Sharpe ratio means better risk-adjusted performance (more return for each unit of volatility). It indicates how efficiently a strategy converts risk into return and is defined as:
	\begin{equation}
		SR = \frac{R_i-R_f}{\sigma_i}
	\end{equation}
	where $R_f$ is the risk-free rate of return, $R_i$ is the average investment return, and $\sigma_i$ is the standard deviation of the investment returns.
	
	\item \textbf{Maximum Drawdown} (MDD): Maximum drawdown is the largest peak-to-trough decline observed in the investment value over the entire period. Specifically, it is the greatest percentage loss from a peak to a subsequent trough before the investment value recovers to a new high. It is a critical indicator of downside risk for a trading strategy and reflects its vulnerability to deep slumps. It is defined as follows:
	\begin{equation}
		MDD = \max_{t \in [1, T]} \left( \frac{V_t - \max_{i \in [1, t]} V_i}{\max_{i \in [1, t]} V_i} \right)
	\end{equation}
	where $V_t$ denotes the investment value at time $t$.
	
	\item \textbf{Calmar Ratio} (CAR): The Calmar ratio is a risk-adjusted performance metric that compares a strategy's annual return to its worst drawdown risk. It highlights how well a trading strategy balances returns against extreme losses and is defined as:
	\begin{equation}
		CAR = \frac{R_i - R_f}{MDD}.
	\end{equation}
	
	\item \textbf{Sortino Ratio} (SOR): The Sortino ratio is a variant of the Sharpe ratio that measures risk-adjusted return focusing only on downside risk. In essence, it is the excess return per unit of downside deviation, indicating how well a strategy compensates for downside volatility. Mathematically,
	\begin{equation}
		SOR = \frac{R_i - R_f}{\sigma_{\text{down}}}
	\end{equation}
	where $\sigma_{\text{down}}$ is the standard deviation of negative returns or those that fall below a threshold.
	
\end{enumerate}

\subsection{Baselines}

A detailed comparison against a broad range of benchmark strategies establishes the robustness of the proposed method. These benchmarks represent a diverse range of approaches, including traditional financial methods, reinforcement learning-based models, and recent studies incorporating auxiliary tasks for stock trading. The baseline trading methods employed in this study are outlined as follows:

\paragraph{Traditional Finance Methods}
\begin{enumerate}
	\item \textbf{Market Index}: The market index is a weighted aggregation used to estimate the performance of a group of stocks of a particular market or sector. It is a measure used to evaluate the performance of a strategy relative to general market trends. In this study, we compared our proposed methodology against DJI for the US market, FTSE for the UK market, Sensex for the Indian market, and TWII for the Taiwan market.
	
	\item \textbf{Buy and Hold}: The buy-and-hold strategy involves buying and holding the assets over the entire investment horizon without active trading. It assumes no rebalancing, and benefits from long-term market appreciation.
	
	\item \textbf{MVO} \cite{markowitz1990foundations}: Mean-Variance Optimization is a robust portfolio construction technique that aims to balance expected return against risk. It relies on estimates of expected returns, covariances, and assumes normally distributed returns, forming a foundational approach in modern portfolio theory.
	
	\item \textbf{Random Trading}: Random trading is a naive baseline where trades are executed randomly without any predictive model or signal. It functions as a lower bound to evaluate the effectiveness of trading strategies by distinguishing gains resulting from skillful decision-making rather than random chance.
	
\end{enumerate}

\paragraph{Reinforcement Learning Methods} 
\begin{enumerate}
	\item \textbf{PPO} \cite{schulman2017proximal}: The conventional Proximal Policy Optimization (PPO) model explicitly adapted for stock trading.
	
	\item \textbf{Volatility-Scaled Deep RL} (VS-DRL)\cite{zhang2019deep}: This study incorporates volatility scaling into the reward function to dynamically adjust position sizes based on market volatility, thereby enhancing robustness and risk management.
	
	\item \textbf{Sharpe Ratio Reward Shaping} (SRRS) \cite{rodinos2023sharpe}: This work implements a reward shaping technique by directly integrating an approximation of the Sharpe ratio into the reward function alongside profit and loss. 
	
	\item \textbf{FinRL DDPG} \cite{liu2021finrl}: The Deep Deterministic Policy Gradient (DDPG) model, implemented via the FinRL library, customized for stock trading.
	\item \textbf{FinRL SAC} \cite{liu2021finrl}: The Soft Actor-Critic (SAC) model for stock trading, implemented using the FinRL framework.
	
	\item \textbf{Adaptive} \cite{yang2020deep}: An ensemble trading strategy combining three actor-critic algorithms: A2C, DDPG, and PPO, to leverage their complementary strengths.
	
	\item \textbf{DREB} \cite{orra2024dynamic}: A dynamic ensemble method that uses Bayesian optimization to assign time-varying weights to multiple DRL agents for automated stock trading.
\end{enumerate}

\paragraph{Auxiliary Task Methods}
\begin{enumerate}
	\item \textbf{Policy-Aware Auxiliary Task Model} (PA-AXT) \cite{konstantinos2022reinforcement}: This work presents variations of a DRL trading agent with auxiliary regression tasks predicting next reward or next close percentage change.
	
	\item \textbf{PPO-AXT} \cite{arabha2024improving}: This model extends PPO by integrating an auxiliary task that performs unsupervised clustering of processed market features via Autoencoder and K-Means.
	
	\item \textbf{DeepScalper} \cite{sun2022deepscalper}: DeepScalper is a DRL framework designed for intraday trading with a large action space. It incorporates a risk-aware auxiliary task that predicts market volatility to guide the agent in balancing profit maximization with risk control.
\end{enumerate}

\subsection{Experiment Settings}

All the experimental studies are performed under consistent environmental settings to ensure model comparability. To simulate realistic trading dynamics, the trading agent is allotted an initial capital of $1,000,000$ ($1$ million) at the beginning of the trading period. The agent incurs a transaction fee of $0.1\%$ of the overall transaction on both buy and sell orders. All stock transactions are executed at the day's closing price. Bayesian optimization \cite{snoek2012practical} is utilized to tune the hyperparameters of the proposed approach and the benchmark DRL models. We employed the Hyperopt library, a Python-based optimization framework, to efficiently conduct the hyperparameter tuning process. The hyperparameter ranges are selected based on insights from empirical studies \cite{mnih2016asynchronous, schulman2017proximal}. The complete set of tuned hyperparameters for the proposed model, together with their respective search ranges, is reported in Table \ref{tab:hyperparameter}.

\begin{table}[!htp]\centering
	\caption{Hyperparameter configuration and search space for the proposed QUESTrader model.}\label{tab:hyperparameter}
		\begin{tabular}{ccc}\toprule
			\textbf{Hyperparameter} &\textbf{Range} \\\midrule
			Hidden Dimension &[2,512] \\
			Number of Layers &[1,8] \\
			Activation Function &[ReLU, Tanh, Sigmoid] \\
			Learning Rate &[$e^{-8}$, $e^{-1}$] \\
			Dropout &[0,0.5] \\
			Gamma &[0.9,0.99] \\
			PPO Epochs &[5,50] \\
			Value Coefficient &[0.01,0.5] \\
			Auxiliary Coefficient &[0.01,0.5] \\
			Entropy Coefficient &[0.01,0.1] \\
			Unroll Length &[1,2,4,10,20,50] \\
			GVF Questions &[2,4,8,16,32,64,128] \\
			\bottomrule
		\end{tabular}
	\end{table}

\section{Results and Discussion \label{Results}}

This section presents a comprehensive assessment of QUESTrader against all baselines across the four markets. It includes a comparative performance evaluation against baseline methods, visualization studies of cumulative returns, action trajectories, and risk profiles, as well as an ablation study analyzing the impact of key hyperparameters.

\subsection{Performance Evaluation}

The effectiveness of the proposed QUESTrader model is assessed by benchmarking its performance against the aforementioned baseline methods across four diverse datasets. The performance of these models is evaluated using a range of risk and return metrics, including Returns, Sharpe ratio, Maximum drawdown, Calmar ratio, and Sortino ratio. A detailed performance comparison of the models on the DJI, FTSE, Sensex, and TAIEX datasets is presented in Tables ~\ref{tab:dow}--\ref{tab:twse}, respectively. Each DRL-based model is trained over five independent runs, and the mean values of the evaluation metrics along with their corresponding standard deviations are reported. The best-performing results are highlighted in bold.

\begin{table}[!htp]\centering
	\caption{Performance comparison of QUESTrader and the baseline methods on the DJI dataset.}\label{tab:dow}
	\resizebox{\textwidth}{!}{ 
		\begin{tabular}{lccccccc}\toprule
			\textbf{Models} &\textbf{Annual return (\%)} &\textbf{Cumulative return (\%)} &\textbf{Sharpe Ratio} &\textbf{Max drawdown (\%)} &\textbf{Calmar ratio} &\textbf{Sortino Ratio} \\\midrule
			Buy-Hold &12.857 &16.097 &0.884 &10.433 &1.054 &1.066 \\
			DJI &8.234 &10.258 &0.724 &9.331 &0.882 &1.040 \\
			MVO \cite{markowits1952portfolio} &11.744 &14.687 &1.247 &8.365 &2.188 &1.796 \\
			Random &0.982 &1.213 &0.650 &1.759 &0.834 &1.524 \\
			VS-DRL \cite{zhang2019deep} &14.548 $\pm$ 1.42 &18.250 $\pm$ 1.76 &0.993 $\pm$ 0.08 &13.232 $\pm$ 0.91 &1.099 $\pm$ 0.10 &1.407 $\pm$ 0.12 \\
			SRRS \cite{rodinos2023sharpe} &16.069 $\pm$ 1.53 &20.190 $\pm$ 1.89 &1.424 $\pm$ 0.11 &9.774 $\pm$ 0.72 &2.371 $\pm$ 0.18 &2.003 $\pm$ 0.16 \\
			FinRL DDPG \cite{liu2021finrl} &10.132 $\pm$ 1.08 &12.649 $\pm$ 1.21 &0.757 $\pm$ 0.07 &10.988 $\pm$ 0.88 &0.922 $\pm$ 0.09 &1.071 $\pm$ 0.11 \\
			FinRL SAC \cite{liu2021finrl} &13.023 $\pm$ 1.31 &16.309 $\pm$ 1.54 &0.899 $\pm$ 0.08 &11.459 $\pm$ 0.94 &1.136 $\pm$ 0.11 &1.269 $\pm$ 0.12 \\
			Adaptive \cite{yang2020deep} &13.922 $\pm$ 1.26 &17.452 $\pm$ 1.61 &1.232 $\pm$ 0.10 &\textbf{7.138 $\pm$ 0.58} &1.950 $\pm$ 0.16 &1.781 $\pm$ 0.14 \\
			DREB \cite{orra2024dynamic} &17.562 $\pm$ 1.59 &22.101 $\pm$ 2.02 &1.359 $\pm$ 0.11 &9.342 $\pm$ 0.74 &1.879 $\pm$ 0.15 &2.018 $\pm$ 0.17 \\
			PA-AXT \cite{konstantinos2022reinforcement} &16.026 $\pm$ 1.48 &20.135 $\pm$ 1.86 &1.055 $\pm$ 0.09 &12.308 $\pm$ 0.93 &1.938 $\pm$ 0.16 &1.604 $\pm$ 0.13 \\
			PPO-AXT \cite{arabha2024improving} &17.937 $\pm$ 1.61 &22.581 $\pm$ 2.07 &1.226 $\pm$ 0.10 &12.004 $\pm$ 0.89 &1.494 $\pm$ 0.13 &1.803 $\pm$ 0.15 \\
			Deep Scalper \cite{sun2022deepscalper} &18.176 $\pm$ 1.67 &22.889 $\pm$ 2.14 &1.215 $\pm$ 0.10 &10.469 $\pm$ 0.81 &2.325 $\pm$ 0.18 &1.908 $\pm$ 0.15 \\
			PPO \cite{schulman2017proximal} &15.674 $\pm$ 1.44 &19.685 $\pm$ 1.81 &1.076 $\pm$ 0.09 &11.713 $\pm$ 0.90 &1.338 $\pm$ 0.12 &1.575 $\pm$ 0.13 \\
			QUESTrader† &\textbf{21.785 $\pm$ 1.42} &\textbf{27.536 $\pm$ 1.91} &\textbf{1.459 $\pm$ 0.09} &10.369 $\pm$ 0.68 &\textbf{2.394 $\pm$ 0.14} &\textbf{2.143 $\pm$ 0.13} \\
			\bottomrule
	\end{tabular}}
\end{table}

For the DJI dataset, QUESTrader delivers the strongest overall performance. It posts an annual return of $21.785\%$ and a cumulative return of $27.536\%$, comfortably ahead of the best baselines. DeepScalper reaches $18.176\%$ annual return, PPO‑AXT is at $17.937\%$, and DREB achieves $17.562\%$. The Sharpe ratio achieved by the proposed model is $1.459$, the highest among all methods. The closest competitors are SRRS with $1.424$ and DREB with $1.359$, while MVO records $1.247$. In terms of downside risk, QUESTrader records a maximum drawdown of $10.369\%$. While this is not the lowest—Adaptive reports $7.138\%$ and MVO $8.365\%$—the drawdown remains well controlled compared to higher-return peers such as VS-DRL ($13.232\%$). This balance between return and risk is reflected in a Calmar ratio of $2.394$ and a Sortino ratio of $2.143$, both representing the best performance across all methods. The closest competitors are SRRS with a Calmar ratio of $2.371$ and DREB with a Sortino ratio of $2.018$. The pattern suggests that the auxiliary‑question representations help the PPO learner convert short‑horizon signals into stable gains, while avoiding the sharp equity dips seen in several baselines. It is also notable that the simple PPO baseline yields $15.674\%$ annual return and $1.076$ Sharpe, which underscores the contribution of the learned auxiliaries beyond standard policy optimization. Overall, the DJI results show a decisive improvement in risk‑adjusted as well as absolute performance.

\begin{table}[!htp]\centering
	\caption{Performance comparison of QUESTrader and the baseline methods on the FTSE dataset.}\label{tab:ftse}
	\resizebox{\textwidth}{!}{ 
		\begin{tabular}{lccccccc}\toprule
			\textbf{Models} &\textbf{Annual return (\%)} &\textbf{Cumulative return (\%)} &\textbf{Sharpe Ratio} &\textbf{Max drawdown (\%)} &\textbf{Calmar ratio} &\textbf{Sortino Ratio} \\\midrule
			Buy-Hold &5.651 &6.984 &0.774 &11.081 &0.816 &0.906  \\
			FTSE &9.828 &12.140 &0.833 &10.183 &0.896 &1.117  \\
			MVO \cite{markowits1952portfolio} &8.429 &10.609 &0.938 &\textbf{10.062} &0.972 &1.071  \\
			Random &-0.240 &-0.301 &-0.117 &1.701 &-0.141 &-0.165  \\
			VS-DRL \cite{zhang2019deep} &12.750 $\pm$ 1.15 &16.184 $\pm$ 1.47 &0.791 $\pm$ 0.07 &14.289 $\pm$ 1.05 &0.892 $\pm$ 0.08 &1.248 $\pm$ 0.11  \\
			SRRS \cite{rodinos2023sharpe} &10.797 $\pm$ 1.02 &13.674 $\pm$ 1.29 &0.436 $\pm$ 0.05 &30.851 $\pm$ 2.01 &0.349 $\pm$ 0.04 &0.778 $\pm$ 0.09  \\
			FinRL DDPG \cite{liu2021finrl} &10.878 $\pm$ 1.03 &12.892 $\pm$ 1.21 &0.554 $\pm$ 0.05 &13.178 $\pm$ 0.98 &0.761 $\pm$ 0.07 &0.868 $\pm$ 0.09  \\
			FinRL SAC \cite{liu2021finrl} &11.847 $\pm$ 1.10 &15.022 $\pm$ 1.39 &0.940 $\pm$ 0.08 &17.423 $\pm$ 1.22 &1.121 $\pm$ 0.10 &1.285 $\pm$ 0.11  \\
			Adaptive \cite{yang2020deep} &12.322 $\pm$ 1.12 &15.633 $\pm$ 1.44 &0.803 $\pm$ 0.07 &12.889 $\pm$ 0.97 &0.956 $\pm$ 0.09 &1.167 $\pm$ 0.10  \\
			DREB \cite{orra2024dynamic} &13.916 $\pm$ 1.21 &17.688 $\pm$ 1.56 &0.647 $\pm$ 0.06 &14.029 $\pm$ 1.03 &0.991 $\pm$ 0.09 &1.014 $\pm$ 0.10 \\
			PA-AXT \cite{konstantinos2022reinforcement} &11.378 $\pm$ 1.06 &14.467 $\pm$ 1.34 &0.705 $\pm$ 0.06 &16.078 $\pm$ 1.16 &0.708 $\pm$ 0.07 &1.181 $\pm$ 0.11 \\
			PPO-AXT \cite{arabha2024improving} &12.191 $\pm$ 1.10 &15.553 $\pm$ 1.42 &0.722 $\pm$ 0.06 &17.862 $\pm$ 1.25 &0.684 $\pm$ 0.07 &1.263 $\pm$ 0.11  \\
			Deep Scalper \cite{sun2022deepscalper} &14.336 $\pm$ 1.25 &18.592 $\pm$ 1.64 &0.686 $\pm$ 0.06 &17.995 $\pm$ 1.28 &0.773 $\pm$ 0.07 &1.254 $\pm$ 0.11 \\
			PPO \cite{schulman2017proximal} &13.018 $\pm$ 1.18 &16.586 $\pm$ 1.52 &0.630 $\pm$ 0.06 &17.814 $\pm$ 1.24 &0.731 $\pm$ 0.07 &1.053 $\pm$ 0.10  \\
			QUESTrader† &\textbf{19.164 $\pm$ 1.36} &\textbf{24.596 $\pm$ 1.79} &\textbf{1.124 $\pm$ 0.08} &12.165 $\pm$ 0.92 &\textbf{1.575 $\pm$ 0.12} &\textbf{1.774 $\pm$ 0.13}  \\
			\bottomrule
	\end{tabular}}
\end{table}

On FTSE, the gains are again broad‑based. QUESTrader achieves $19.164\%$ annual and $24.596\%$ cumulative return, while the next best in absolute returns are DeepScalper at $14.336\%$ and DREB at $13.916\%$. The Sharpe ratio rises to $1.124$, clearly above strong baselines such as FinRL‑SAC ($0.940$) and MVO ($0.938$). The maximum drawdown is $12.165\%$, which is competitive though not the lowest—MVO achieves $10.062\%$ but at the cost of substantially lower returns. Crucially, the Calmar ratio is $1.575$ and the Sortino ratio is $1.774$, both of which are the highest among the competitors. Methods that aggressively chase return, such as DREB and DeepScalper, show larger drawdowns ($14.029\%$ and $17.995\%$) and weaker Calmar values ($0.991$ and $0.773$). Even the PPO‑AXT variant, which uses handcrafted auxiliary tasks, remains behind on all three risk‑adjusted metrics. The results indicate that the discovered questions generalize to the FTSE regime with only slight performance degradation, while also contributing to the stability of the policy head under this volatile market. Hence, the method advances both return maximization and drawdown control, without relying on fixed predefined auxiliary heuristics.

\begin{table}[!htp]\centering
	\caption{Performance comparison of QUESTrader and the baseline methods on the Sensex dataset.}\label{tab:sensex}
	\resizebox{\textwidth}{!}{ 
		\begin{tabular}{lccccccc}\toprule
			\textbf{Models} &\textbf{Annual return (\%)} &\textbf{Cumulative return (\%)} &\textbf{Sharpe Ratio} &\textbf{Max drawdown (\%)} &\textbf{Calmar ratio} &\textbf{Sortino Ratio} \\\midrule
			Buy-Hold &5.461 &6.768 &0.482 &15.721 &0.411 &0.614 \\
			Sensex &6.663 &8.175 &0.538 &15.757 &0.422 &0.741 \\
			MVO \cite{markowits1952portfolio} &7.830 &9.652 &0.703 &18.076 &0.433 &1.034 \\
			Random &-0.605 &-0.739 &-0.254 &3.990 &-0.151 &-0.343 \\
			VS-DRL \cite{zhang2019deep} &11.065 $\pm$ 1.04 &13.686 $\pm$ 1.29 &0.723 $\pm$ 0.06 &17.156 $\pm$ 1.21 &0.644 $\pm$ 0.06 &1.252 $\pm$ 0.11 \\
			SRRS \cite{rodinos2023sharpe} &10.688 $\pm$ 0.99 &13.214 $\pm$ 1.25 &0.709 $\pm$ 0.06 &10.650 $\pm$ 0.84 &1.003 $\pm$ 0.09 &1.050 $\pm$ 0.10 \\
			FinRL DDPG \cite{liu2021finrl} &7.679 $\pm$ 0.82 &9.465 $\pm$ 0.93 &0.527 $\pm$ 0.05 &12.028 $\pm$ 0.96 &0.638 $\pm$ 0.06 &0.729 $\pm$ 0.08 \\
			FinRL SAC \cite{liu2021finrl} &9.203 $\pm$ 0.91 &11.361 $\pm$ 1.06 &0.647 $\pm$ 0.06 &10.851 $\pm$ 0.87 &0.850 $\pm$ 0.08 &0.929 $\pm$ 0.09 \\
			Adaptive \cite{yang2020deep} &11.578 $\pm$ 1.06 &14.829 $\pm$ 1.36 &0.799 $\pm$ 0.07 &15.412 $\pm$ 1.14 &0.806 $\pm$ 0.07 &1.126 $\pm$ 0.10 \\
			DREB \cite{orra2024dynamic} &15.138 $\pm$ 1.31 &18.802 $\pm$ 1.63 &0.916 $\pm$ 0.07 &13.089 $\pm$ 0.97 &1.156 $\pm$ 0.10 &1.341 $\pm$ 0.11 \\
			PA-AXT \cite{konstantinos2022reinforcement} &12.351 $\pm$ 1.12 &15.297 $\pm$ 1.41 &0.860 $\pm$ 0.07 &16.356 $\pm$ 1.18 &0.755 $\pm$ 0.07 &1.236 $\pm$ 0.11 \\
			PPO-AXT \cite{arabha2024improving} &13.583 $\pm$ 1.19 &16.844 $\pm$ 1.49 &0.870 $\pm$ 0.07 &16.420 $\pm$ 1.20 &0.827 $\pm$ 0.08 &1.228 $\pm$ 0.11 \\
			Deep Scalper \cite{sun2022deepscalper} &13.423 $\pm$ 1.18 &16.642 $\pm$ 1.47 &0.856 $\pm$ 0.07 &17.783 $\pm$ 1.27 &0.754 $\pm$ 0.07 &1.221 $\pm$ 0.11 \\
			PPO \cite{schulman2017proximal} &11.948 $\pm$ 1.08 &14.791 $\pm$ 1.37 &0.768 $\pm$ 0.06 &10.653 $\pm$ 0.85 &1.121 $\pm$ 0.10 &1.129 $\pm$ 0.10 \\
			QUESTrader† &\textbf{16.727 $\pm$ 1.21} &\textbf{20.809 $\pm$ 1.54} &\textbf{1.003 $\pm$ 0.07} &\textbf{10.584 $\pm$ 0.79} &\textbf{1.263 $\pm$ 0.10} &\textbf{1.488 $\pm$ 0.11} \\
			\bottomrule
	\end{tabular}}
\end{table}

The Sensex dataset yields comparatively conservative returns across all methods, which makes risk-adjusted metrics particularly revealing. QUESTrader records $16.727\%$ annual and $20.809\%$ cumulative return, improving over DREB at 15.138\%, PPO‑AXT at $13.583\%$, and DeepScalper at $13.423\%$. The Sharpe ratio attains a value of $1.003$, the highest among all methods, with the following best being $0.916$ achieved by DREB. Drawdown is also well managed at $10.584\%$, comparable to SRRS ($10.650\%$) and PPO ($10.653\%$), and much lower than those for DeepScalper ($17.783\%$) and Adaptive ($15.412\%$). Consequently, the method achieves a Calmar ratio of $1.263$ and a Sortino ratio of $1.488$, both representing the best performance in the table. These outcomes are significant because Sensex exhibits abrupt shifts in volatility and liquidity. The discovered GVF questions provide informative multi-scale features that support PPO in maintaining performance stability during market shifts. The improvement over the plain PPO baseline ($11.948\%$ annual return, Sharpe ratio $0.768$) further highlights that the auxiliary-task mechanism is primarily responsible for the performance gains.

\begin{table}[!htp]\centering
	\caption{Performance comparison of QUESTrader and the baseline methods on the TAIEX dataset.}\label{tab:twse}
	\resizebox{\textwidth}{!}{ 
		\begin{tabular}{lccccccc}\toprule
			\textbf{Models} &\textbf{Annual return (\%)} &\textbf{Cumulative return (\%)} &\textbf{Sharpe Ratio} &\textbf{Max drawdown (\%)} &\textbf{Calmar ratio} &\textbf{Sortino Ratio} \\\midrule
			Buy-Hold &15.937 &19.053 &0.874 &15.097 &0.910 &1.184 \\
			TWII &17.554 &20.991 &0.893 &18.692 &0.939 &1.173 \\
			MVO \cite{markowits1952portfolio} &18.007 &21.468 &1.738 &\textbf{8.065} &2.173 &2.171 \\
			Random &1.778 &2.089 &0.797 &2.198 &0.810 &1.066 \\
			VS-DRL \cite{zhang2019deep} &25.987 $\pm$ 2.21 &31.173 $\pm$ 2.64 &1.038 $\pm$ 0.09 &21.581 $\pm$ 1.42 &1.204 $\pm$ 0.11 &1.519 $\pm$ 0.13 \\
			SRRS \cite{rodinos2023sharpe} &23.797 $\pm$ 2.05 &28.498 $\pm$ 2.41 &0.906 $\pm$ 0.08 &24.492 $\pm$ 1.61 &0.971 $\pm$ 0.09 &1.276 $\pm$ 0.11 \\
			FinRL DDPG \cite{liu2021finrl} &21.345 $\pm$ 1.94 &25.514 $\pm$ 2.17 &0.917 $\pm$ 0.08 &19.898 $\pm$ 1.35 &1.072 $\pm$ 0.10 &1.325 $\pm$ 0.12 \\
			FinRL SAC \cite{liu2021finrl} &19.582 $\pm$ 1.78 &23.375 $\pm$ 2.01 &0.940 $\pm$ 0.08 &17.423 $\pm$ 1.22 &1.129 $\pm$ 0.10 &1.285 $\pm$ 0.11 \\
			Adaptive \cite{yang2020deep} &23.856 $\pm$ 2.06 &28.571 $\pm$ 2.39 &1.158 $\pm$ 0.09 &14.653 $\pm$ 1.04 &1.628 $\pm$ 0.13 &1.633 $\pm$ 0.13 \\
			DREB \cite{orra2024dynamic} &25.776 $\pm$ 2.19 &30.915 $\pm$ 2.58 &1.134 $\pm$ 0.09 &15.317 $\pm$ 1.08 &1.688 $\pm$ 0.14 &1.708 $\pm$ 0.14 \\
			PA-AXT \cite{konstantinos2022reinforcement} &24.516 $\pm$ 2.11 &29.374 $\pm$ 2.47 &1.138 $\pm$ 0.09 &16.267 $\pm$ 1.15 &1.507 $\pm$ 0.13 &1.624 $\pm$ 0.13 \\
			PPO-AXT \cite{arabha2024improving} &21.141 $\pm$ 1.92 &25.269 $\pm$ 2.16 &0.972 $\pm$ 0.08 &17.101 $\pm$ 1.19 &1.236 $\pm$ 0.11 &1.384 $\pm$ 0.12 \\
			Deep Scalper \cite{sun2022deepscalper} &26.833 $\pm$ 2.25 &32.740 $\pm$ 2.71 &1.236 $\pm$ 0.10 &16.231 $\pm$ 1.14 &1.653 $\pm$ 0.14 &1.860 $\pm$ 0.15 \\
			PPO \cite{schulman2017proximal} &21.513 $\pm$ 1.96 &25.718 $\pm$ 2.19 &1.001 $\pm$ 0.08 &19.525 $\pm$ 1.33 &1.102 $\pm$ 0.10 &1.490 $\pm$ 0.12 \\
			QUESTrader† &\textbf{30.279 $\pm$ 2.01} &\textbf{38.603 $\pm$ 2.58} &\textbf{1.803 $\pm$ 0.11} &13.632 $\pm$ 0.96 &\textbf{2.559 $\pm$ 0.16} &\textbf{2.310 $\pm$ 0.15} \\
			\bottomrule
	\end{tabular}}
\end{table}

TAIEX represents a high-yield environment in our study, and QUESTrader adapts effectively to this regime. The model posts $30.279\%$ annual and $38.603\%$ cumulative return, well above DeepScalper ($26.833\%$), VS‑DRL ($25.987\%$), DREB ($25.776\%$), and PA‑AXT ($24.516\%$). With a Sharpe ratio of $1.803$, the proposed method achieves the best risk-adjusted performance, outperforming MVO ($1.738$), which nevertheless exhibits considerably lower returns. The maximum drawdown is $13.632\%$, which is higher than MVO's $8.065\%$ but substantially lower than deep RL baselines such as PPO ($19.525\%$) and VS-DRL ($21.581\%$). Unlike MVO, which minimizes drawdown at the expense of returns, QUESTrader achieves a more favorable balance between return generation and risk control. Driven by substantially higher returns, the Calmar ratio rises to $2.559$ and the Sortino ratio to $2.310$, both leading all methods. These values exceed the next best results from DeepScalper ($1.653$ Calmar; $1.860$ Sortino) and DREB ($1.688$ Calmar; $1.708$ Sortino), underscoring QUESTrader's ability to balance profitability with downside risk. The combination suggests that the auxiliary questions discovered on‑policy provide a richer pool of predictive targets than hand‑crafted designs, allowing the policy to scale up exposure when the market trend is reliable and to scale down when risk rises. The four datasets present a consistent picture. QUESTrader tops annual return, cumulative return, Sharpe, Calmar, and Sortino on every dataset, while keeping max drawdown competitive. The discovered auxiliary questions give PPO a better state representation and more reliable advantage signals, leading to steadier policy improvement. 

\subsection{Visualization Study}

\begin{figure}[ht]
	\centering
	\includegraphics[scale=0.3]{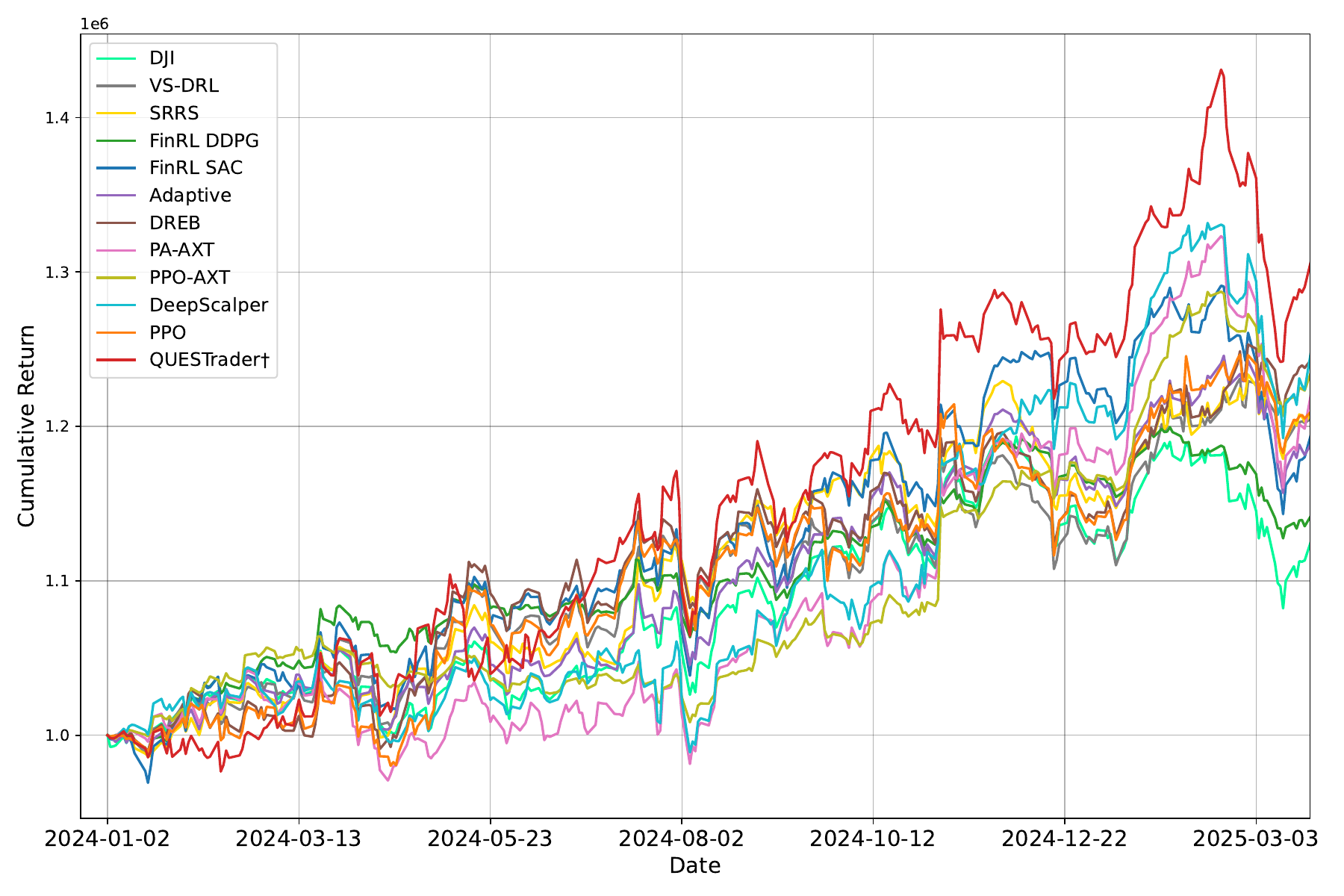}
	\caption{Cumulative return trajectories of the proposed QUESTrader and baseline models on the DJI dataset.}
	\label{fig:dow}
\end{figure}

\begin{figure}[ht]
	\centering
	\includegraphics[scale=0.3]{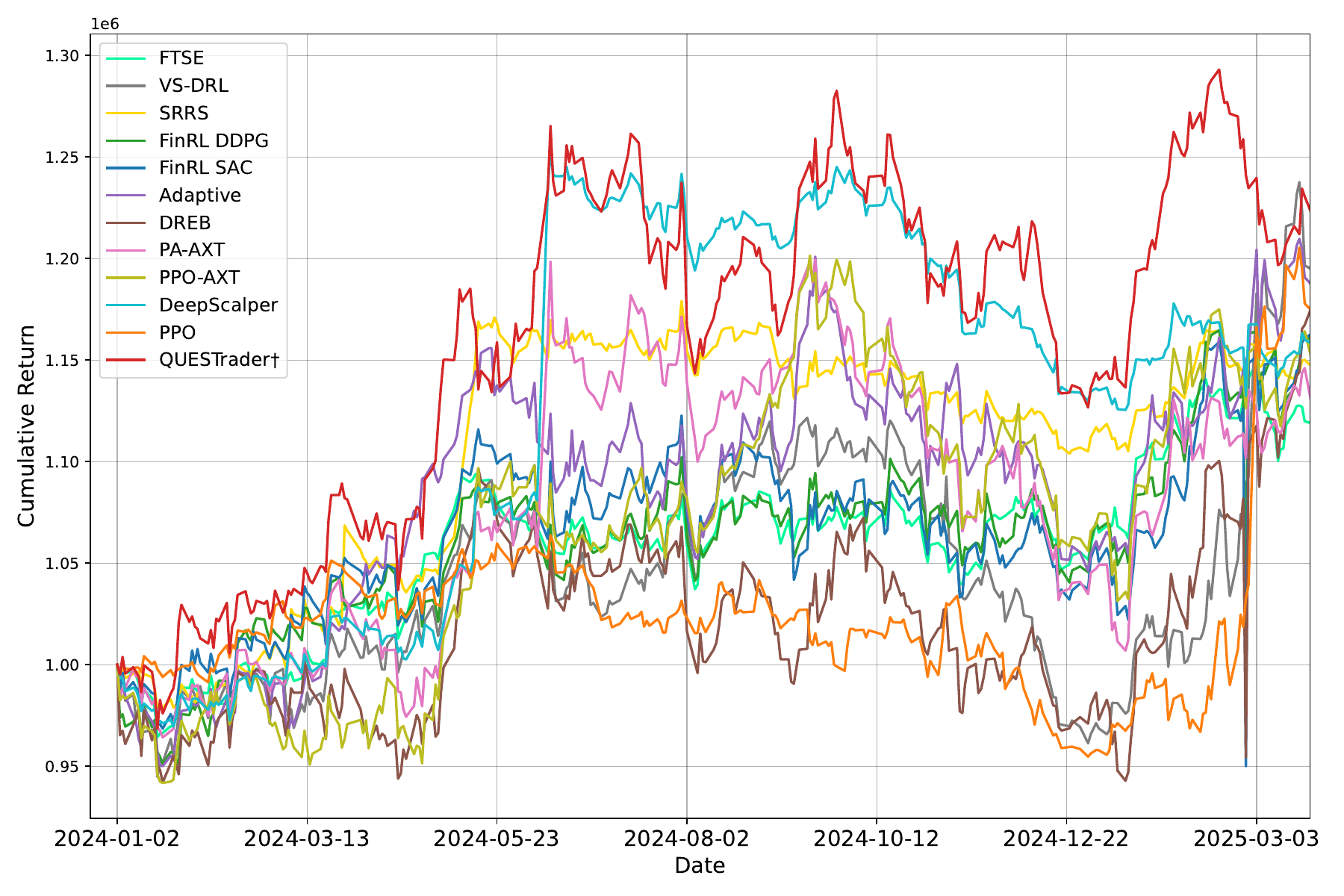}
	\caption{Cumulative return trajectories of the proposed QUESTrader and baseline models on the FTSE dataset.}
	\label{fig:ftse}
\end{figure}

\begin{figure}[ht]
	\centering
	\includegraphics[scale=0.3]{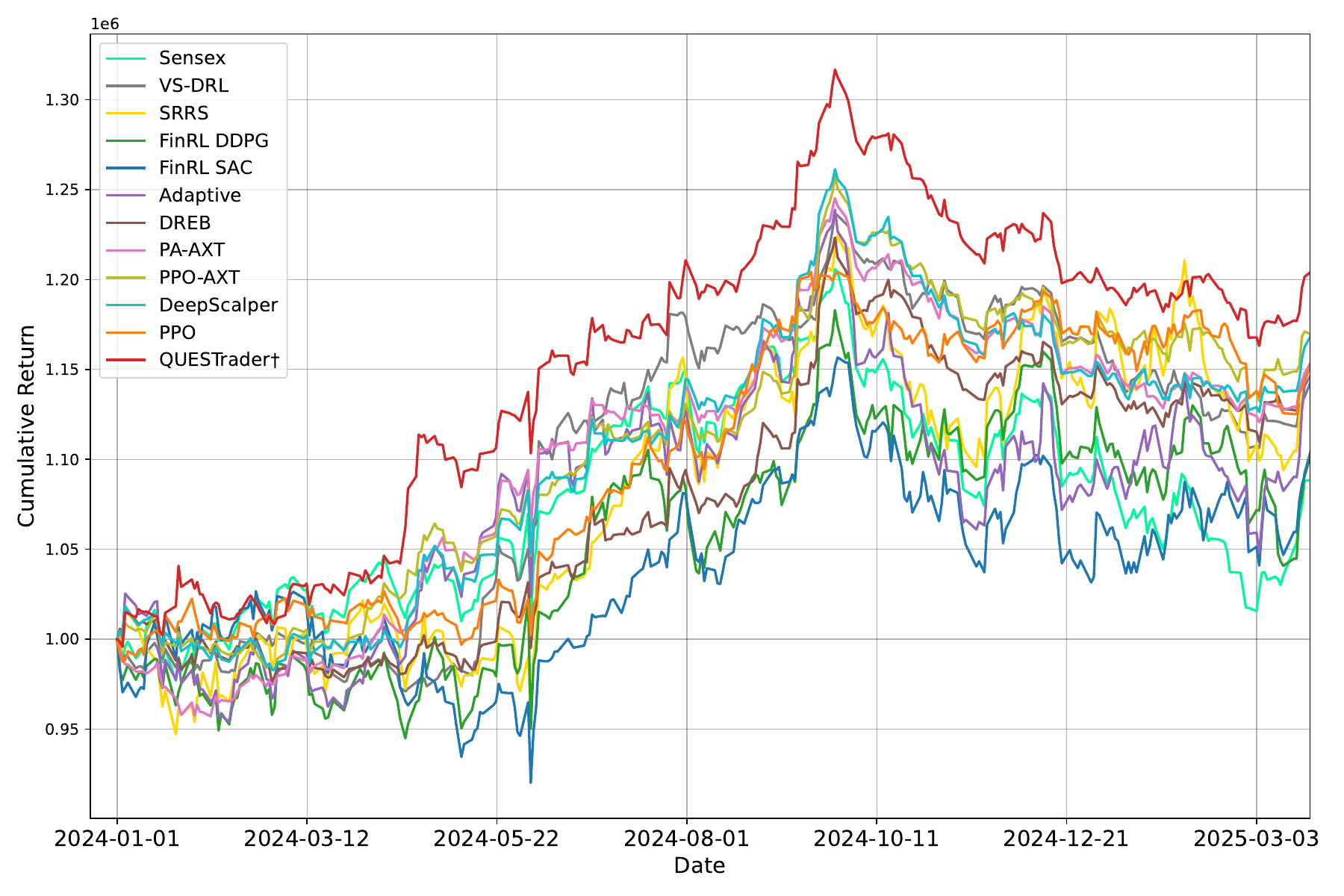}
	\caption{Cumulative return trajectories of the proposed QUESTrader and baseline models on the Sensex dataset.}
	\label{fig:sensex}
\end{figure}

\begin{figure}[ht]
	\centering
	\includegraphics[scale=0.3]{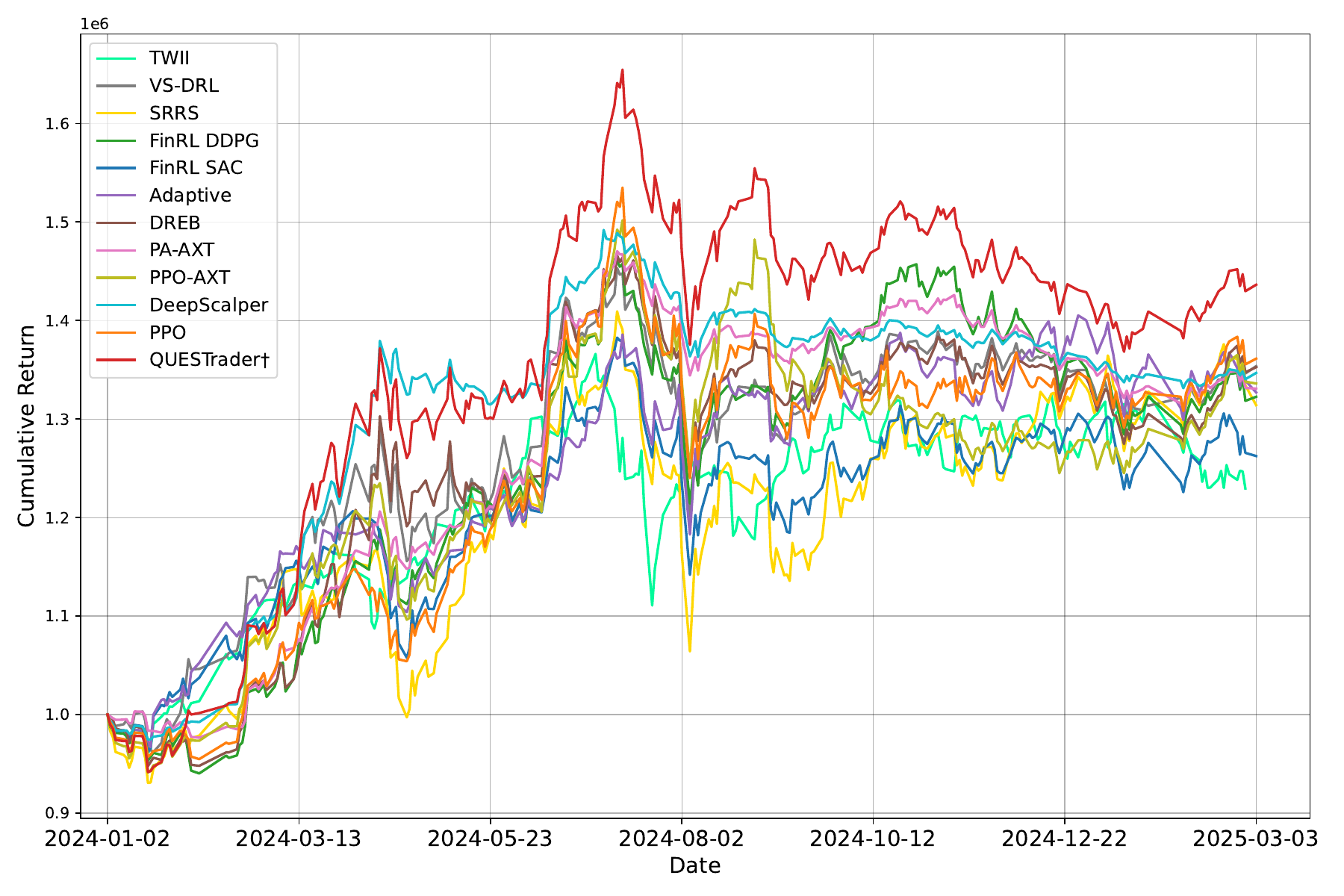}
	\caption{Cumulative return trajectories of the proposed QUESTrader and baseline models on the TAIEX dataset.}
	\label{fig:twse}
\end{figure}

We performed a visual comparison of cumulative return trajectories for the proposed QUESTrader model and the baseline methods. Figures~\ref{fig:dow}--\ref{fig:twse} present the cumulative return plots of the models on the DJI, FTSE, Sensex, and TAIEX datasets, respectively. On the DJI dataset, the equity curve of QUESTrader rises early, experiences a mid-period soft patch with a shallow dip, and then resumes with a steady upward trajectory. Competing RL methods recover later and show multiple small drawdowns around the same region. That difference in time in the market creates a visible gap by the end. On FTSE, many baselines experience a long plateau after an initial rise. Their curves oscillate around a horizontal range, showing only modest upward movement near the end of the window. QUESTrader also pauses during this interval but resumes its upward trajectory earlier and closes with a clear margin over the baselines. This pattern indicates enhanced regime detection and a more efficient use of low-volatility periods to recalibrate risk. On the Sensex dataset, all curves exhibit slower growth, which is characteristic of a more conservative regime. Here, the gains primarily arise from effective loss avoidance. The proposed method avoids several false starts that hinder the baselines, with its equity curve advancing in steady, incremental steps and exhibiting only short-lived dips. On TAIEX, the environment is faster and more momentum‑driven. The curves of aggressive baselines show sharp surges but relinquish substantial gains during reversals. QUESTrader's curve is steeper during upswings and tighter during drawdowns, with narrower troughs and faster rebounds. This pattern indicates that the model scales exposure during persistent trends and reduces it swiftly when signals weaken. The proposed discovery of auxiliary questions improves timing discipline, risk control, and durable compounding. 

\begin{figure}[ht!]
	\centering
	\begin{subfigure}[b]{0.45\textwidth}
		\centering
		\includegraphics[width=\textwidth]{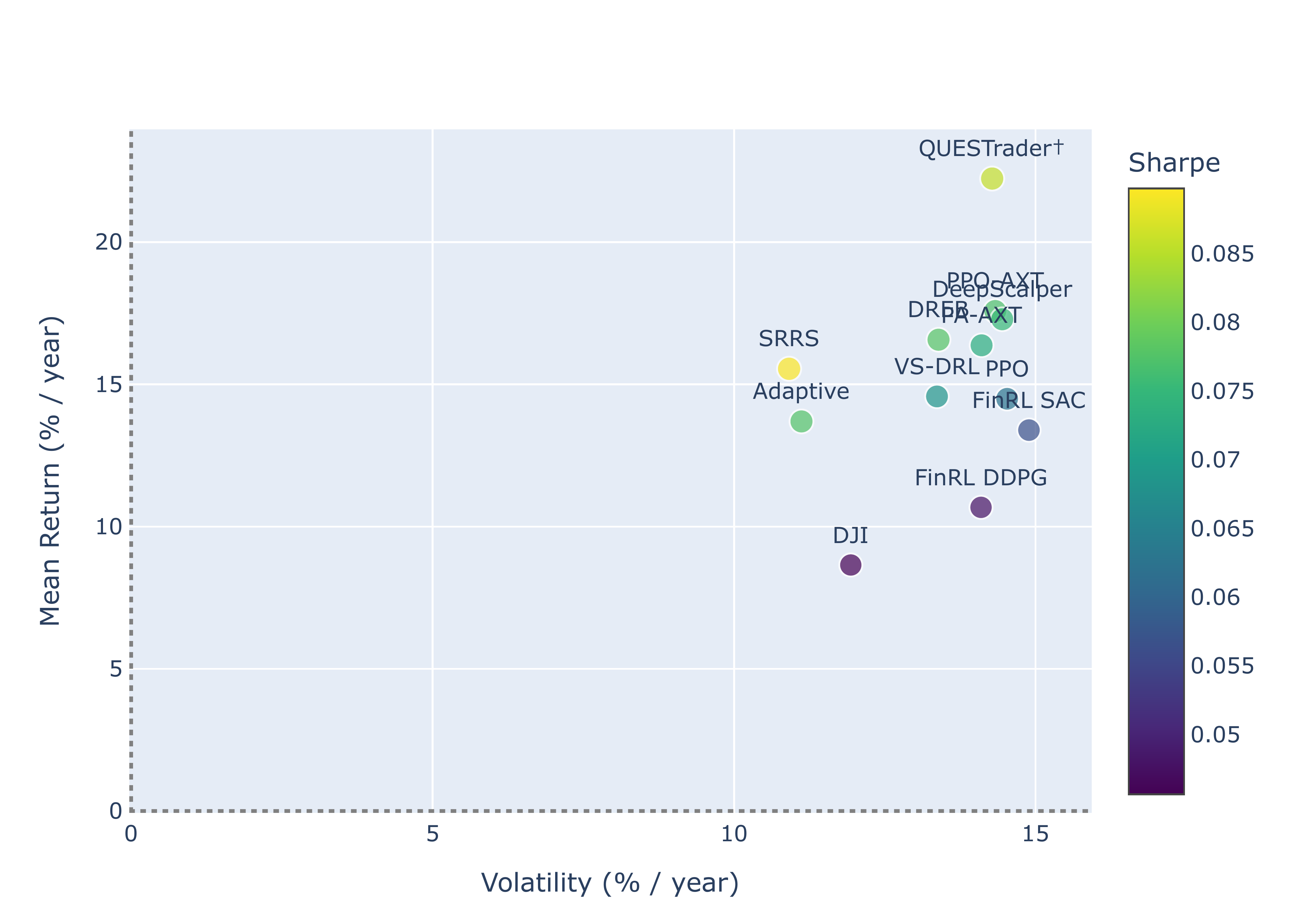}
		\caption{DJI}
	\end{subfigure}
	\hfill
	\begin{subfigure}[b]{0.45\textwidth}
		\centering
		\includegraphics[width=\textwidth]{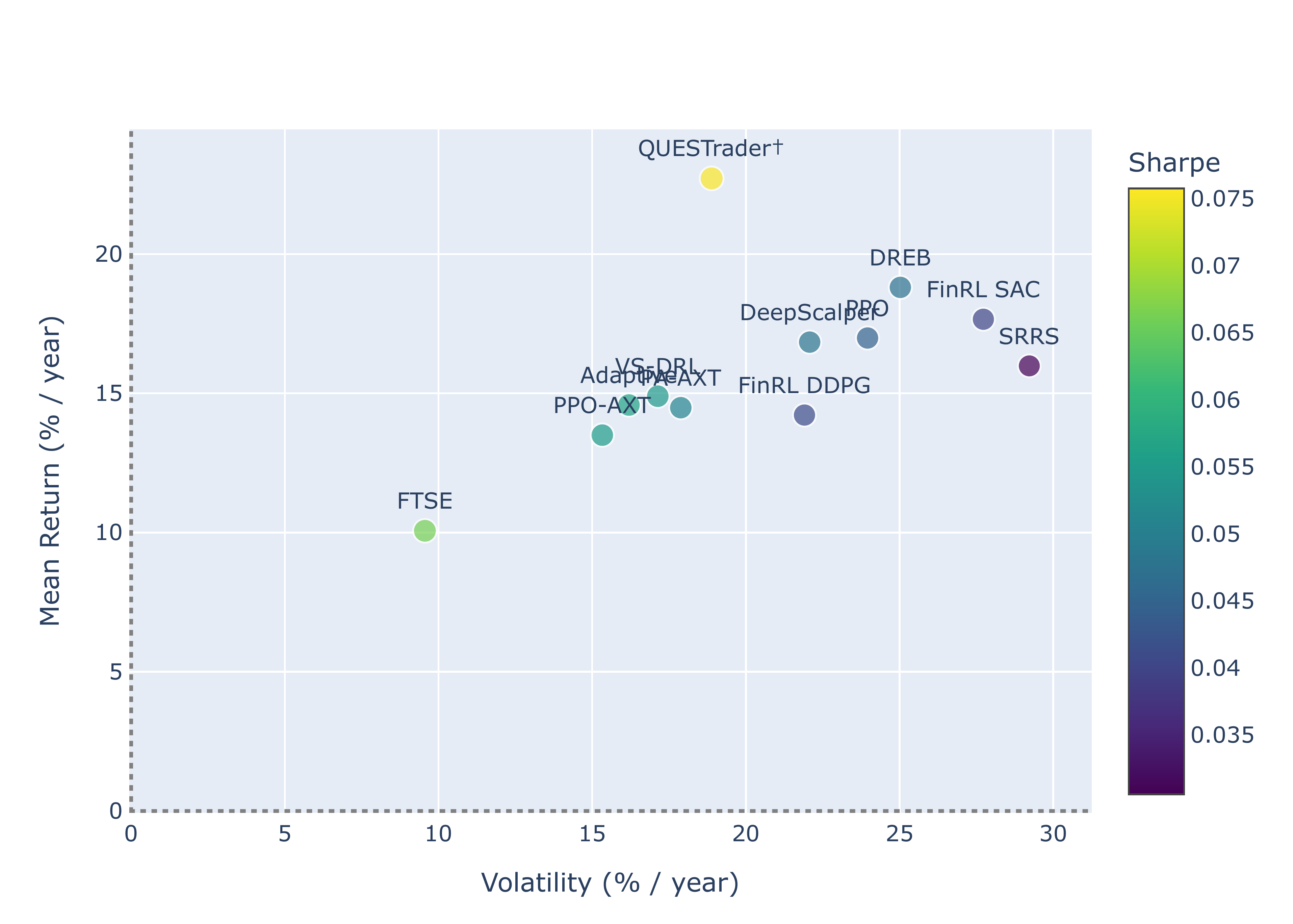}
		\caption{FTSE}
	\end{subfigure}
	
	\begin{subfigure}[b]{0.45\textwidth}
		\centering
		\includegraphics[width=\textwidth]{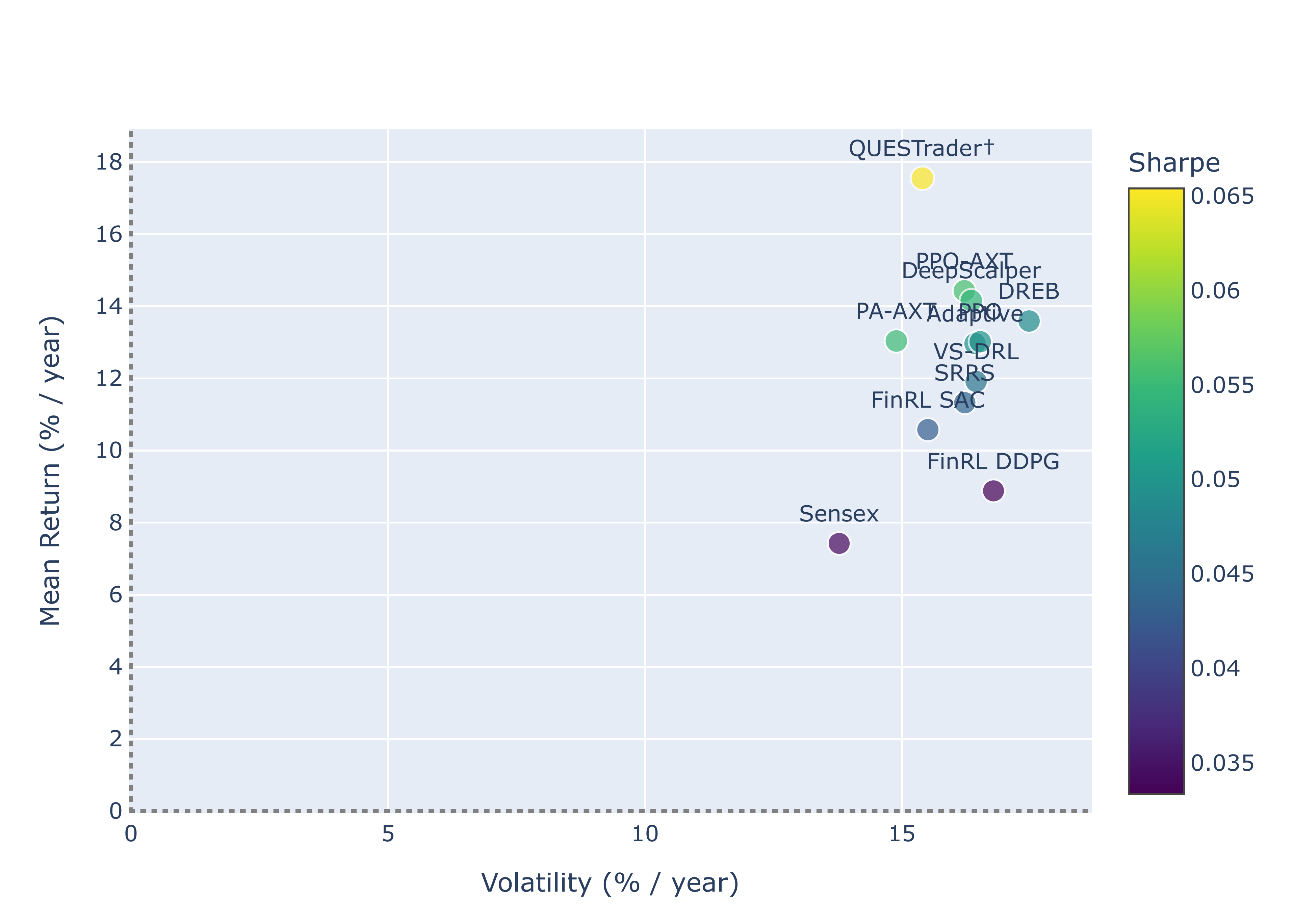}
		\caption{Sensex}
	\end{subfigure}
	\hfill
	\begin{subfigure}[b]{0.45\textwidth}
		\centering
		\includegraphics[width=\textwidth]{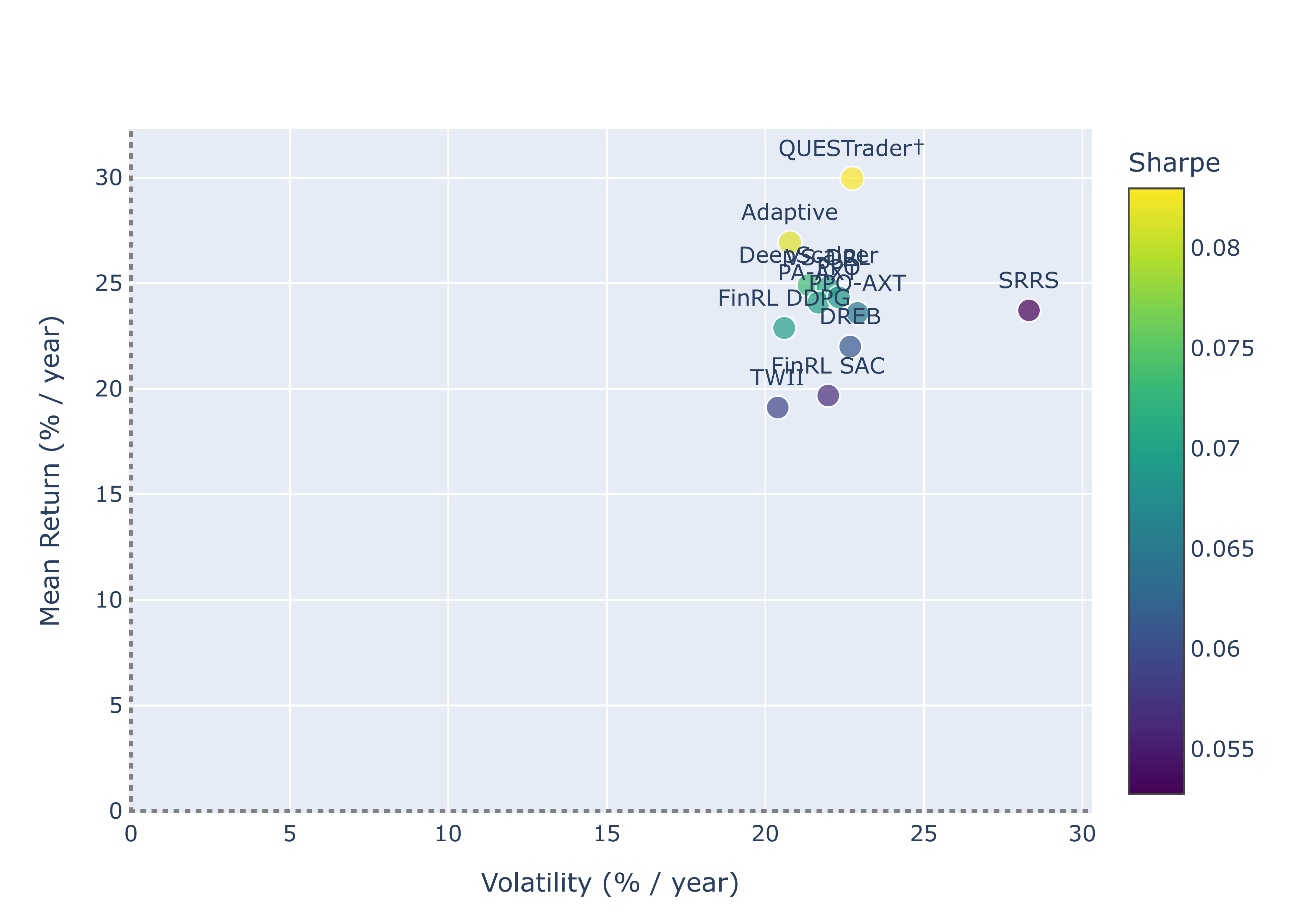}
		\caption{TAIEX}
	\end{subfigure}
	\caption{Risk–return scatter plot across all methods and markets.}
	\label{fig:risk}
\end{figure}

Figure \ref{fig:risk} displays, for each dataset, a scatter of models in mean–variance space: the x-axis is annualized volatility (risk), the y-axis is annualized mean return, and the colour encodes the Sharpe ratio. This visual is important because it places every method on the same risk–reward canvas. On DJI, the baseline cluster sits roughly between $10-15\%$ volatility and $10-18\%$ return. Methods such as SRRS, DREB, and DeepScalper inch upwards but remain inside the same risk band. QUESTrader lifts itself above the cluster with clearly higher mean return while holding volatility in the same corridor. The colour shading indicates the highest Sharpe on the panel. This tells us that the gains are not merely due to taking larger positions; rather, the auxiliary questions sharpen the state so that PPO chooses its moments better. On FTSE, the picture repeats with a slight twist. Several deep RL baselines push rightwards to $20-30\%$ volatility for modest increases in return, and their Sharpe colours fade accordingly. QUESTrader appears in the upper‑middle of the panel: return near the top, volatility around the mid‑teens, and a brighter Sharpe than its neighbours. This means the method converts noisy trends into profits without amplifying dispersion. On Sensex, returns across methods are lower, and the spread in volatility is narrower. Many baselines sit in a tight cluster around mid‑teens volatility with low double-digit returns; their colours are muted, reflecting average Sharpe. QUESTrader stands above this group with a clean increase in mean return at a similar volatility level, and the colour bar signals the best Sharpe on the panel. This is encouraging because Sensex exhibits abrupt liquidity and policy shocks. On TAIEX, the regime is more momentum-driven, and volatilities are higher overall. Some baselines chase the trend by sliding rightwards to the high‑volatility zone, gaining return but losing Sharpe. QUESTrader sits highest in return with volatility in the same band as competitive baselines, yet its colour remains distinctly brighter. Taken together, these plots therefore confirm that our approach improves the risk‑adjusted frontier and shows robustness across regimes.

\begin{figure}[ht!]
	\centering
	\begin{subfigure}[b]{0.45\textwidth}
		\centering
		\includegraphics[width=\textwidth, height=4cm]{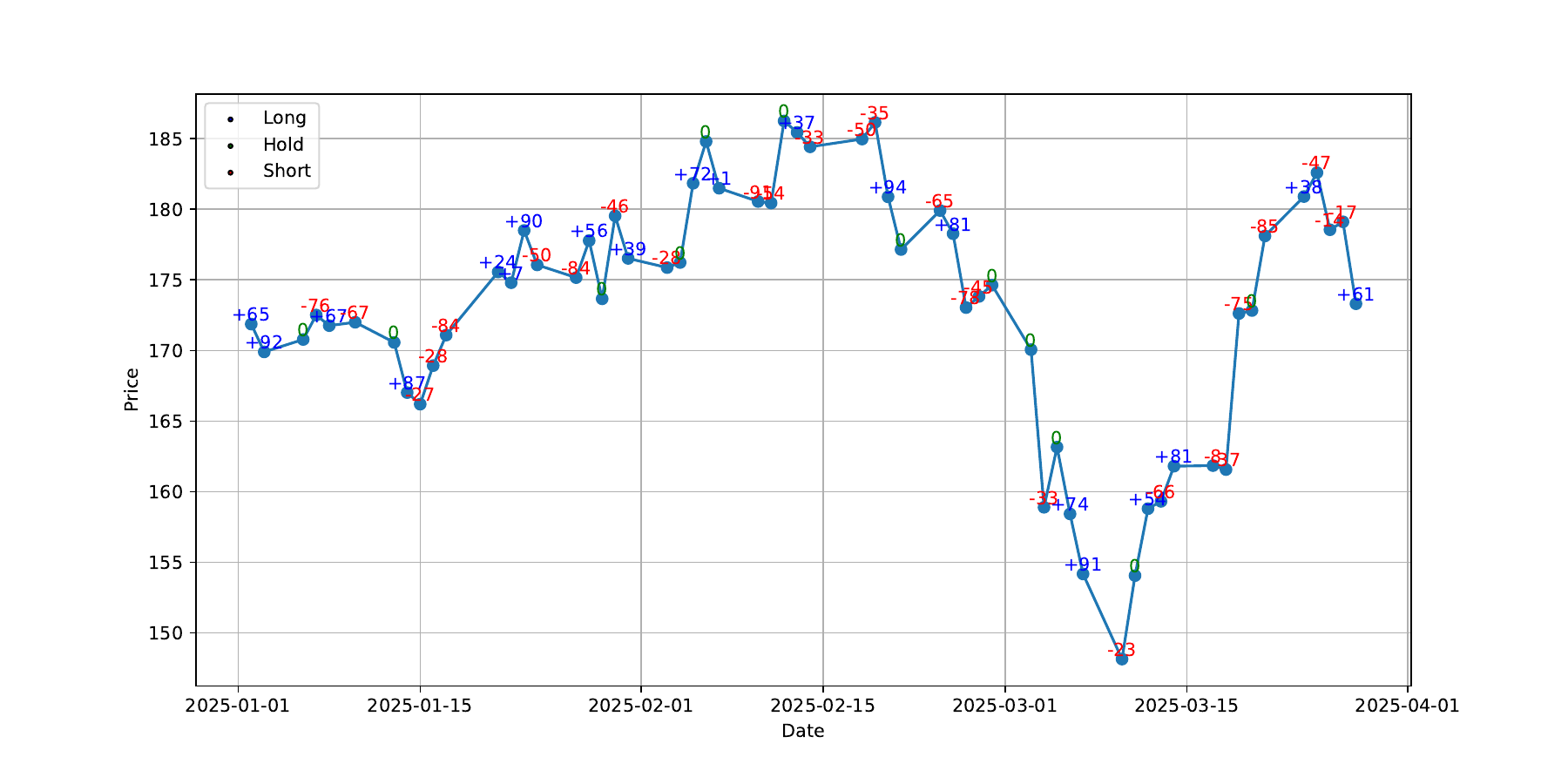}
		\caption{PPO}
	\end{subfigure}
	\hfill
	\begin{subfigure}[b]{0.45\textwidth}
		\centering
		\includegraphics[width=\textwidth, height=4cm]{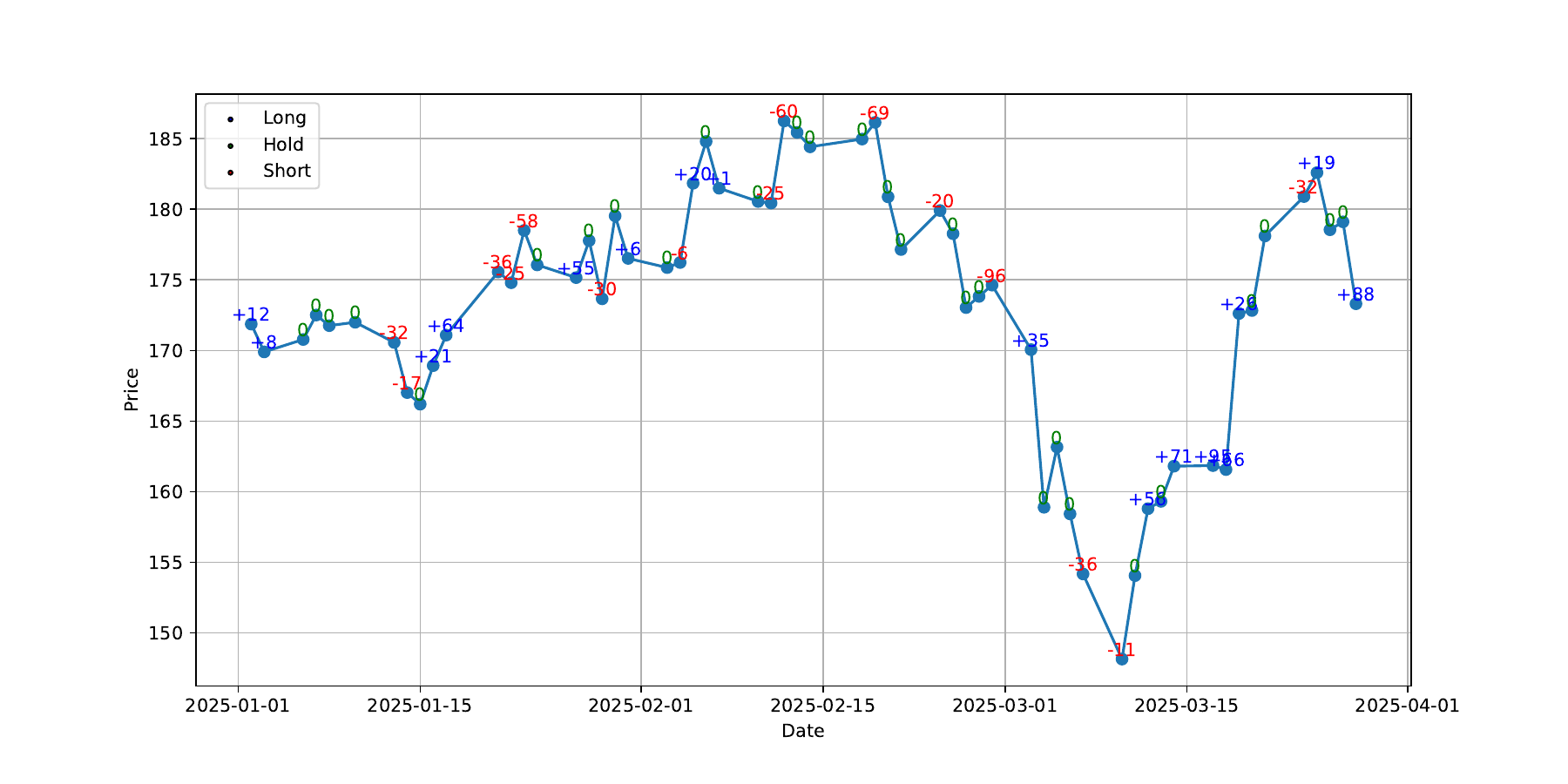}
		\caption{QUESTrader}
	\end{subfigure}
	\caption{Action timelines: PPO baseline versus QUESTrader. Long/Short/Hold decisions overlaid on price. QUESTrader shows fewer whipsaws, earlier regime turns, and more disciplined holds.}
	\label{fig:actions}
\end{figure}

Figure \ref{fig:actions} offers a side‑by‑side trace of actions taken by the plain PPO baseline and QUESTrader on the same stock and period. Each marker shows a long, short, or hold trading decision superimposed on the price path. This visual is informative because it exposes how a method trades, not just how much it earns. The left panel shows PPO's behaviour, and the right panel shows our model's behaviour under the same conditions. A direct comparison highlights three features. First, QUESTrader issues fewer and more coherent switches between long and short. Second, it aligns its positions closer to the local trend structure. Third, it uses holds to sit out noise pockets, thereby reducing unnecessary turnover and cost. In the early rising segment, QUESTrader maintains long positions for longer stretches. PPO, in contrast, flips around minor pullbacks and re‑enters late. They add slippage without a clear benefit. During the sharp drawdown, QUESTrader keeps the short stance intact for most of the descent. PPO shows intermittent counter‑trend longs that are quickly reversed. Such whipsaws are expensive in a costed setting. When the market rebounds, QUESTrader exits the short briskly and rebuilds a long in a stepwise manner. The long run then persists through the recovery with minimal churn. PPO again shows extra flips near the lows and in the first upswing. That delay leads to missed early gains. Across the whole horizon, QUESTrader's decisions look temporally smoother and regime-aware. 

\subsection{Ablation Study}

\begin{figure}[ht!]
	\centering
	\begin{subfigure}[b]{0.45\textwidth}
		\centering
		\includegraphics[width=\textwidth]{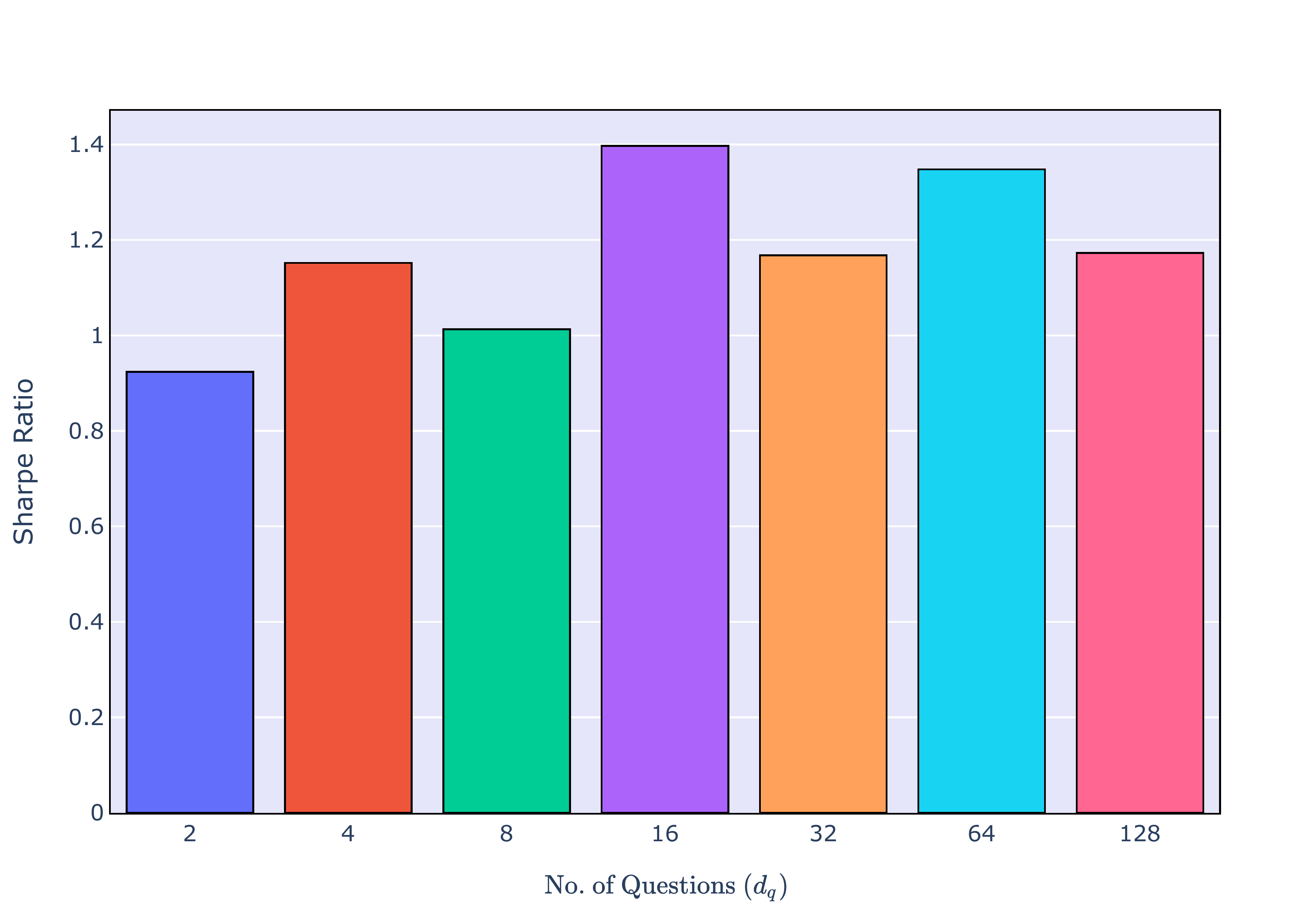}
		\caption{Sharpe ratio vs No. of Questions}
	\end{subfigure}
	\hfill
	\begin{subfigure}[b]{0.45\textwidth}
		\centering
		\includegraphics[width=\textwidth]{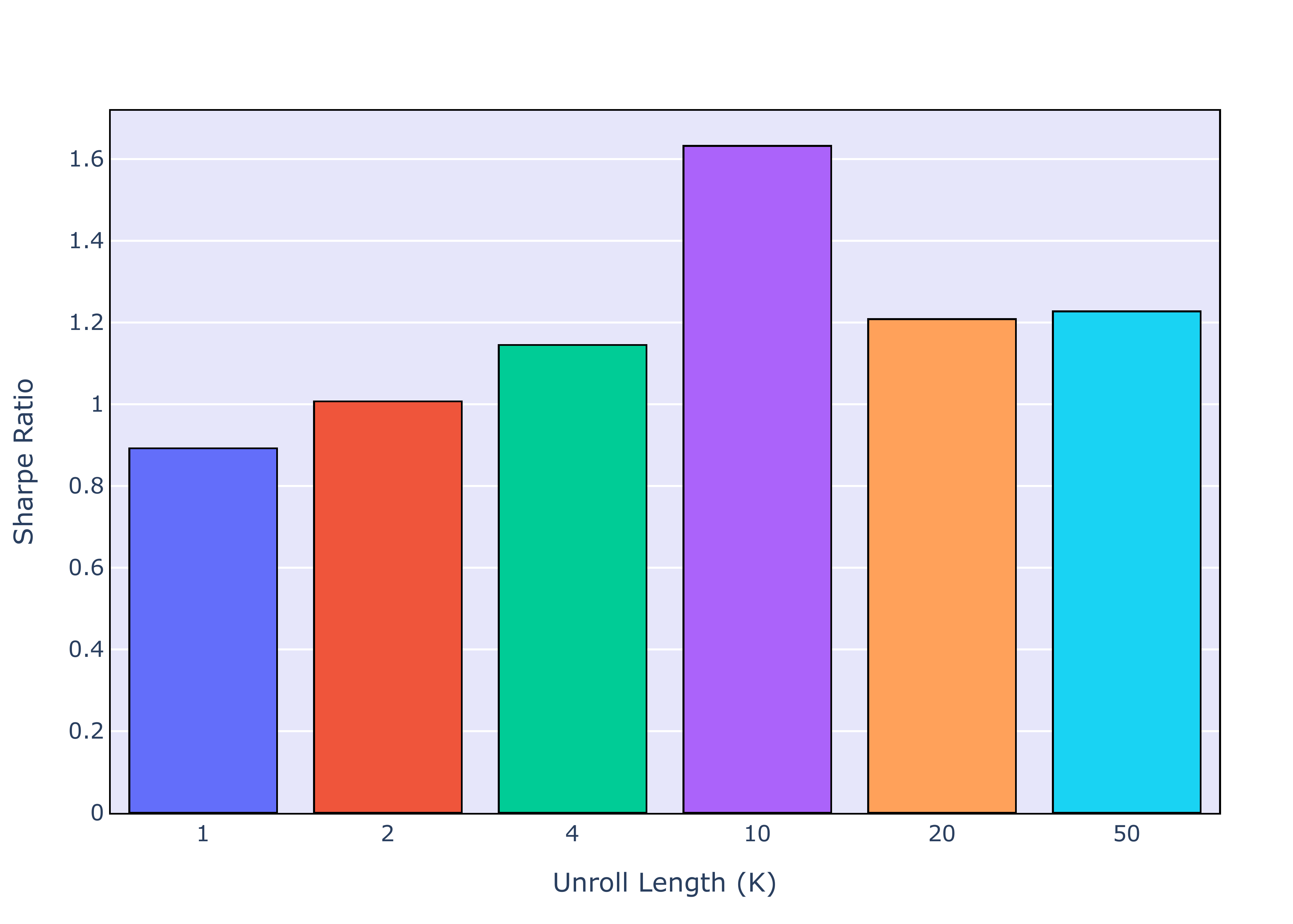}
		\caption{Sharpe ratio vs Unroll Length}
	\end{subfigure}
	\caption{Ablation: effect of number of questions and unroll length on the Sharpe ratio.}
	\label{fig:ablation_sharpe}
\end{figure}

\begin{figure}[ht!]
	\centering
	\begin{subfigure}[b]{0.45\textwidth}
		\centering
		\includegraphics[width=\textwidth]{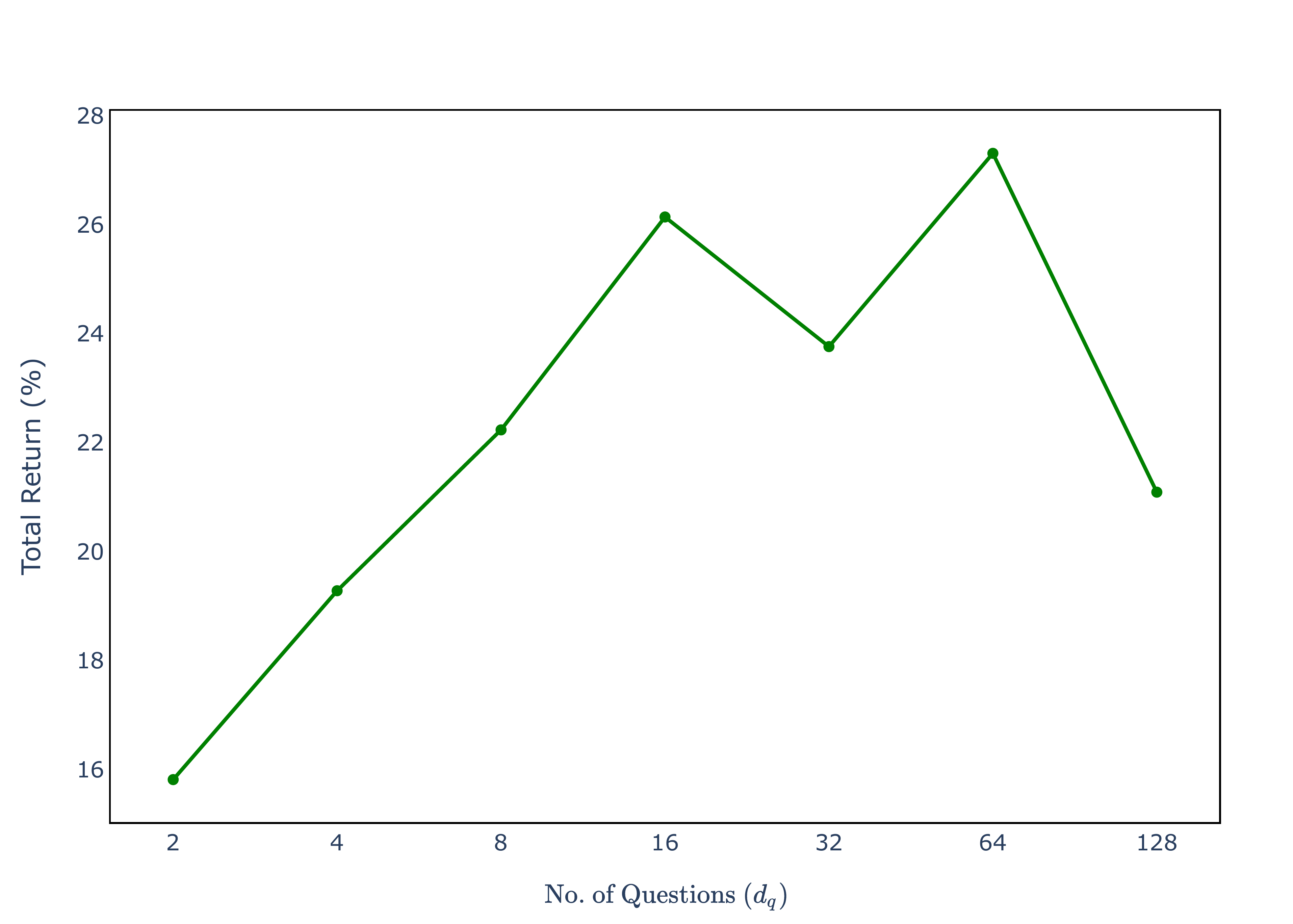}
		\caption{Total return vs No. of Questions}
	\end{subfigure}
	\hfill
	\begin{subfigure}[b]{0.45\textwidth}
		\centering
		\includegraphics[width=\textwidth]{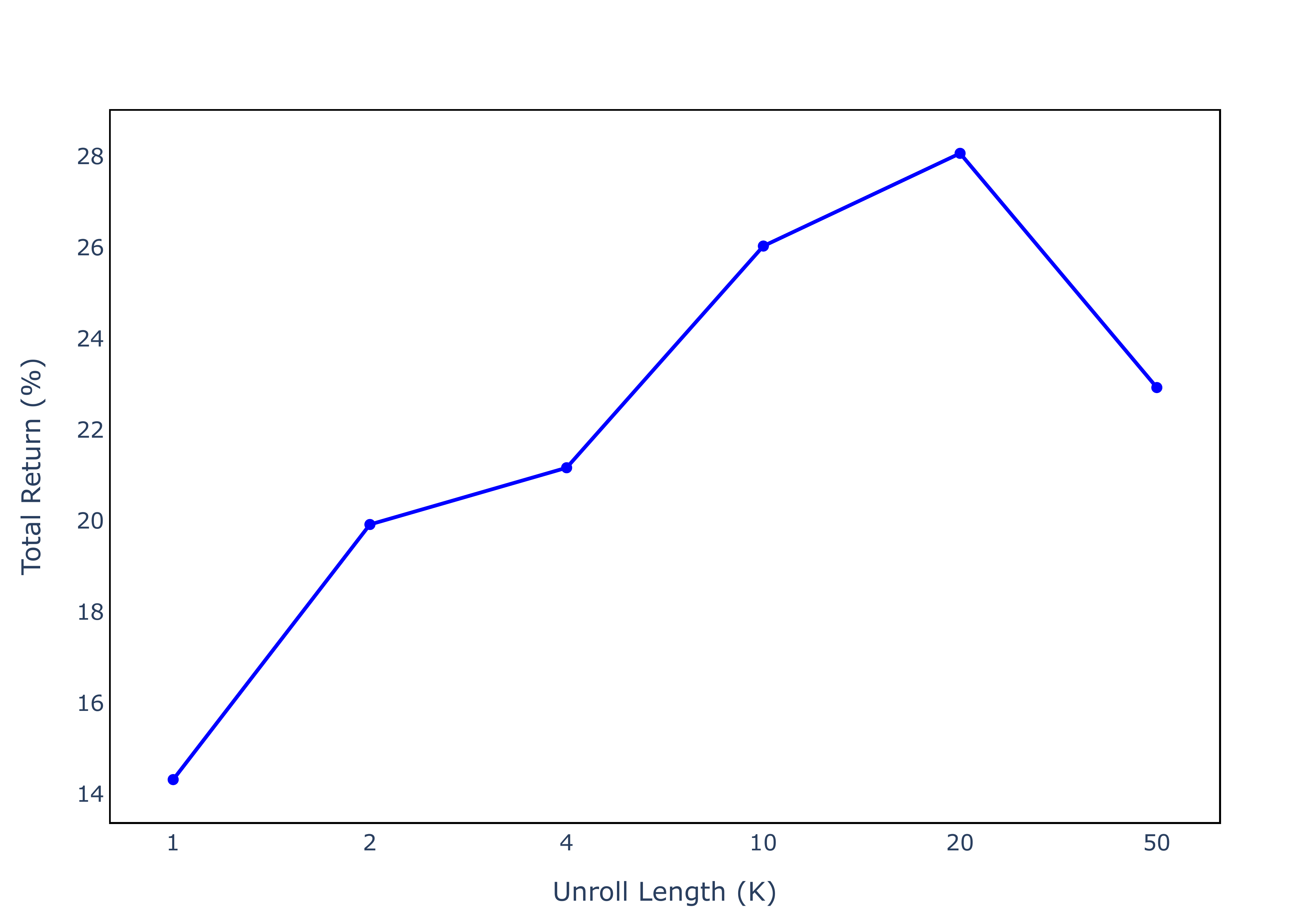}
		\caption{Total return vs Unroll Length}
	\end{subfigure}
	\caption{Ablation: effect of number of questions and unroll length on total return.}
	\label{fig:ablation_return}
\end{figure}

We conduct a focused ablation study on the DJI dataset to investigate how the number of discovered GVF questions ($d_q$) and the inner unroll length ($K$) affect trading performance. These two hyperparameters control, respectively, the breadth of auxiliary predictions and the temporal depth of credit assignment to those predictions within each update cycle. For each configuration, we retrain the full pipeline and evaluate it using the fixed test window. We assess performance using two complementary metrics for each hyperparameter setting: total return (\%) and the Sharpe ratio. Figure \ref{fig:ablation_sharpe} plots the Sharpe ratio against $d_q$ and $K$. And, the Figure \ref{fig:ablation_return} plots the total return against the same two parameters. 

Sharpe improves sharply when we move from very few questions to a moderate bank. With only $d_q=2$, the Sharpe is the lowest ($\approx 0.9$). It rises through $d_q=4$ and $8$, and reaches its peak around $d_q=16$ ($\approx 1.4$). Increasing further to $d_q=32$ yields a mild drop ($\approx 1.2$), while $d_q=64$ recovers to a near‑peak value ($\approx 1.3-1.35$). At $d_q=128$, Sharpe falls again ($\approx 1.15$). The curve suggests a sweet-spot range: too few questions underrepresent market structure. Conversely, too many fragments the learning signal, introduce redundancy, and amplify bootstrap noise in the GVF heads, which reduces risk-adjusted gains. Turning to the unroll length $K$, Sharpe is weak at $K=1$ and $2$ ($\approx 0.9-1.0$), improves at $K=4$ ($\approx 1.15$), and peaks at $K=10$ ($\approx 1.6$). Moving to $K=20$ keeps Sharpe strong ($\approx 1.2$) but below the peak, while $K=50$ is similar ($\approx 1.2$). The pattern matches the intuition behind non‑myopic updates. Very small unrolls cannot reveal the delayed usefulness of discovered questions, yielding limited policy improvement. Moderate unroll lengths propagate credit across several PPO steps and stabilize the representation, thereby enhancing Sharpe performance. In contrast, very long unrolls increase gradient variance and memory demands, which diminishes the marginal benefit. In practice, $d_q \in [16,64]$ with $K \approx 10$ is a safe region if the objective is to maximize Sharpe.

As $d_q$ grows from $2$ to $16$, the return climbs steadily from $~15-16\%$ to $~26\%$, confirming that a richer set of auxiliary questions helps the policy harvest more signal. At $d_q=32$, there is a dip to $~24\%$, suggesting some interference or over‑regularisation of the shared network. Interestingly, $d_q=64$ delivers the highest return ($~27-28\%$), showing that a larger bank can help scale exposure when trends are reliable. Pushing to $d_q=128$ reduces return to $~21\%$, likely due to redundant predictions, slower optimization of many heads, and a weaker representation alignment with the PPO objective. For the unroll length $K$, the return increases monotonically from $~14\%$ at $K=1$ to $~26\%$ at $K=10$ and peaks at $~28\%$ for $K=20$. At $K=50$, the return drops to $~23\%$. The rise from $K=1$ to $K=20$ shows that deeper credit assignment helps the question network propose auxiliaries that lead to better-timed exposure after several updates. The decline at $K=50$ shows diminishing returns, as long unrolls add noise to the meta-gradient and render PPO ratios stale relative to the fixed behavior policy. Taken together, the Sharpe–return pairing suggests a practical operating point where one balances risk‑adjusted quality and absolute gain: $d_q$ around $16-64$ and $K$ around $10-20$.

\section{Conclusion and Future Work \label{Conclusion}}

This work set out to improve reinforcement learning for trading by automatically discovering useful predictions, in contrast to manually designing side objectives. We proposed QUESTrader, a two-network PPO framework that discovers auxiliary questions as GVFs and learns to answer them alongside the trading policy. The Answer/Main network implements the PPO policy alongside the GVF answer heads, whereas the Question network specifies auxiliary questions by emitting per-time cumulants and discount factors. These questions are tuned by a non‑myopic meta‑gradient computed through an inner unroll of PPO updates. The method produces a shared representation that is rich, stable, and sensitive to market scale. Empirical results on DJI, FTSE, Sensex, and TAIEX indicate consistent improvements. The proposed model delivers the highest annual and cumulative returns across all datasets and achieves superior Sharpe, Calmar, and Sortino ratios, while maintaining competitive drawdowns. The cumulative return plots across all four datasets show that the proposed model increases exposure during persistent trends and reduces it rapidly when signals weaken, highlighting its ability to balance profitability with risk control. The risk–return scatter plot places QUESTrader on the north‑west frontier, which means more reward for similar risk. The ablation study clarifies the role of the two key hyperparameters. Performance improves strongly when we move from very few questions to a moderate bank and peaks again near $64$ questions for total return. Performance also rises when the unroll length increases from one to $10-20$ steps and then flattens. Methodologically, the study conveys two key insights. First, auxiliary questions discovered on-policy align more closely with the trading objective than fixed, handcrafted auxiliaries. Second, propagating credit across multiple learner steps is essential for these questions to have a practical impact.

While the present findings are promising, there are several natural directions for extension. This study focuses on on-policy GVFs, but future work could extend the framework to off-policy GVFs and implicit-gradient meta-updates, which may reduce variance and memory overhead. Another promising direction is online regime detection, which would allow the model to adapt both the number of discovered questions ($d_q$) and the unroll length ($K$) dynamically during training. Also, extending the framework to a portfolio setting with cross-sectional questions would enable the capture of spread opportunities and sector-rotation dynamics.

\bibliographystyle{elsarticle-num}
\bibliography{Ref}
	
\end{document}